\documentclass[letterpaper]{article} 
\usepackage{aaai24}  
\usepackage{times}  
\usepackage{helvet}  
\usepackage{courier}  
\usepackage[hyphens]{url}  
\usepackage{graphicx} 
\usepackage{natbib}  
\usepackage{caption} 
\usepackage{algorithm}
\usepackage{algorithmic}

\usepackage{newfloat}
\usepackage{listings}
\DeclareCaptionStyle{ruled}{labelfont=normalfont,labelsep=colon,strut=off} 
\floatstyle{ruled}
\newfloat{listing}{tb}{lst}{}
\floatname{listing}{Listing}
\usepackage{latexsym}
\usepackage{comment}

\usepackage{color} 
\usepackage{multirow} 
\usepackage{amsmath, amsfonts, amsthm, amssymb}
\usepackage{subfigure} 
\usepackage[inline]{enumitem}
\usepackage{booktabs}

\usepackage[switch]{lineno} 

\newtheoremstyle{mysubtask}%
{}{}{\normalfont}{}{\scshape}{.\,}{ }{}
\theoremstyle{mysubtask}

\newtheoremstyle{mytask}%
{}{}{\normalfont}{}{\scshape}{.\,}{ }{}
\theoremstyle{mytask}

\newtheoremstyle{mydef}%
{}{}{\normalfont}{}{\scshape}{.\,}{ }{}
\theoremstyle{mydef}

\usepackage{xcolor}
\usepackage{soul}

\newcommand{\hlc}[2][yellow]{{%
    \colorlet{foo}{#1}%
    \sethlcolor{foo}\hl{#2}}%
}

\definecolor{mygreen}{rgb}{0.286,0.737,0.392}

\usepackage{arydshln}
\usepackage{siunitx}

\date{}

\title{InstructDoc: A Dataset for Zero-Shot Generalization of Visual Document Understanding with Instructions}

\author{
    Ryota Tanaka\textsuperscript{\rm 1,2},
    Taichi Iki\textsuperscript{\rm 1},
    Kyosuke Nishida\textsuperscript{\rm 1}, 
    Kuniko Saito\textsuperscript{\rm 1},
    Jun Suzuki\textsuperscript{\rm 2}
\\
}
\affiliations{
    \textsuperscript{\rm 1}NTT Human Informatics Laboratories, NTT Corporation \\
    \textsuperscript{\rm 2}Tohoku University \\
    \{ryota.tanaka, taichi.iki, kyosuke.nishida, kuniko.saito\}@ntt.com, jun.suzuki@tohoku.ac.jp
}

\begin{document}

\maketitle

\begin{abstract}
We study the problem of completing various visual document understanding (VDU) tasks, e.g., question answering and information extraction, on real-world documents through human-written instructions. To this end, we propose InstructDoc, the first large-scale collection of 30 publicly available VDU datasets, each with diverse instructions in a unified format, which covers a wide range of 12 tasks and includes open document types/formats. Furthermore, to enhance the generalization performance on VDU tasks, we design a new instruction-based document reading and understanding model, InstructDr, that connects document images, image encoders, and large language models (LLMs) through a trainable bridging module. Experiments demonstrate that InstructDr can effectively adapt to new VDU datasets, tasks, and domains via given instructions and outperforms existing multimodal LLMs and ChatGPT without specific training. 
\end{abstract}

\section{Introduction}
Building document artificial intelligence (Document AI) capable of reading and comprehending real-world documents, including webpages, office documents, mobile UIs, etc., has been a long-cherished goal. Toward this goal, numerous works on visual document understanding (VDU) have addressed a wide range of tasks, such as document question answering (QA)~\cite{Mathew_2021_WACV} and information extraction~\cite{jaume2019funsd}. Document data contain both textual and visual objects, with content spread structurally across various locations depending on diverse document types and formats. To address this complexity, previous works have proposed models that aim to improve interactions among text/layout/visual modalities~\cite{xu2020layoutlmv2,appalaraju2021docformer}. However, the diversity of documents and tasks poses a challenge in developing a unified model that can comprehend intricate relationships between text and visual objects across a wide range of document types, formats, and tasks.

To improve the generalizability and adaptivity of unseen vision-language tasks, visual instruction tuning~\cite{xu-etal-2023-multiinstruct,liu2023llava} has been introduced. This approach involves training multimodal large language models (MLLMs) on a collection of images, task inputs, and instructions. However, according to ~\cite{liu2023hidden}, most of the previous visual instruction tuning datasets have primarily focused on understanding visual (non-textual) objects in scene images and existing models struggle with accomplishing tasks that require visual document understanding abilities. While recent works~\cite{zhang2023llavar,ye2023mplugdocowl} attempt to deal with the issue, they still exhibit limitations when generalizing to unseen tasks and documents.

\begin{figure*}[t!]
    \centering
    \includegraphics[width=.99\textwidth]{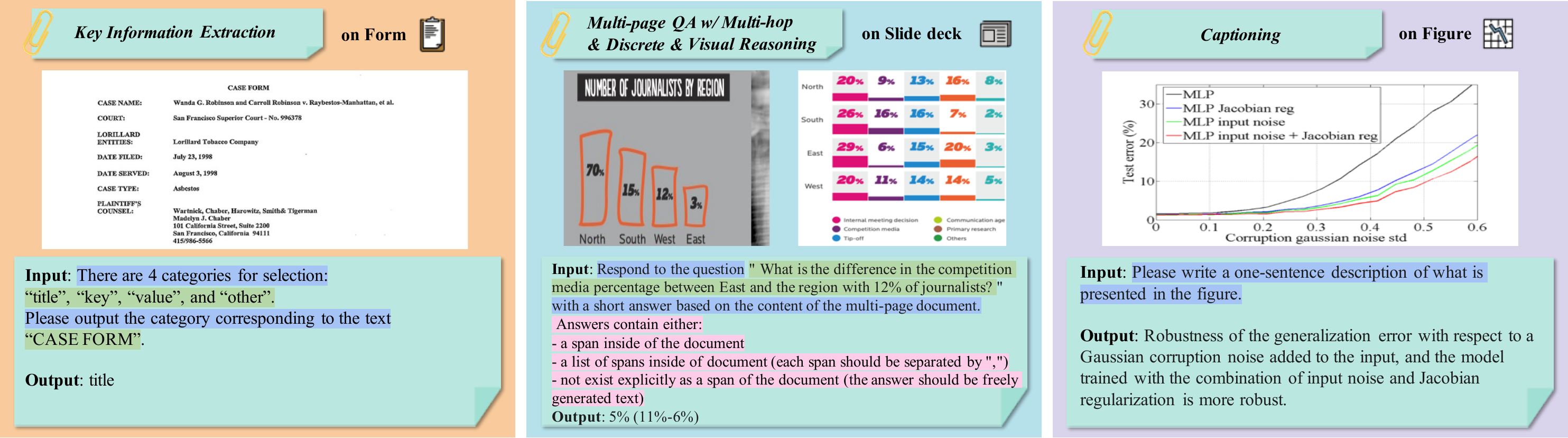}
    \caption{Examples of InstructDoc dataset. The input defines \hlc[blue!30]{\textit{intent}}, \hlc[green!20]{\textit{query and options}}, and \hlc[magenta!30]{\textit{answer style}}. \hlc[green!20]{\textit{query and options}} and outputs are from original datasets. We annotated instructions composed of \hlc[blue!30]{\textit{intent}} and \hlc[magenta!30]{\textit{answer style}} or only \hlc[blue!30]{\textit{intent}}.}
    \label{fig:samples}
\end{figure*}

In this paper, we propose \textbf{InstructDoc}\footnote{Our dataset and codes are publicly available at \url{https://github.com/nttmdlab-nlp/InstructDoc}}, the first large-scale visual instruction tuning dataset that covers a wide range of VDU tasks and datasets (12 diverse tasks created from 30 openly available datasets). Each dataset has diverse instructions annotated by experts, following a unified instruction schema, composed of user's \textit{intent} and \textit{answer style}, for VDU tasks. As shown in Figure~\ref{fig:samples}, 
InstructDoc requires a rich set of abilities, including understanding document layout, visual representations of texts, and relation extraction of objects (e.g., graphs and charts) over open document types/formats with handcrafted instructions. 

Furthermore, to enhance the generalization performance on VDU tasks, we present a \textbf{Instruct}ion-based \textbf{D}ocument \textbf{r}eading and understanding model, InstructDr, which unifies the visual, text, and layout modalities of documents by bridging the gap between a vision encoder and a large language model (LLM) through a new bridging module called Document-former. The Document-former converts documents into a useful feature for the LLM. Experiments show that InstructDr achieves the highest zero-shot performance among existing MLLMs and outperforms ChatGPT on a wide range of VDU datasets with instructions. 

\section{Related Work}
\subsubsection{Visual document understanding.}
Visual documents are ubiquitous and used in diverse applications, including QA on business documents~\cite{Mathew_2021_WACV}, information extraction on receipts~\cite{jaume2019funsd}, and classification over large document collections~\cite{harley2015evaluation}. Due to this diversity, previous works have generally been domain/task-specific, lacking the sharing of underlying data, model architectures, and objectives~\cite{XuLCHWZ20,appalaraju2021docformer,huang2022layoutlmv3}. Although pixel-based methods~\cite{kim2022ocr,lee2023pix2struct} can simplify architectures, these methods have high computational costs (due to the encoding of high-resolution images) and can have degraded performance on new tasks. 
We leverage the reasoning abilities of LLMs and perform all VDU tasks in a unified sequence-to-sequence format with instructions, resulting in improved generalization performance.

\subsubsection{Instruction-following language models.}
Training LLMs with instructions on various NLP tasks has proven effective in improving zero-shot performance of unseen tasks~\cite{wei2021finetuned,iyer2022opt}. Flan~\cite{wei2021finetuned,longpre2023flan}, PromptSource~\cite{bach-etal-2022-promptsource}, and Natural Instructions~\cite{mishra-etal-2022-cross} collected instructions and datasets for a variety of general NLP tasks, such as machine reading comprehension and summarization tasks on plain-text documents. In contrast, we tackle the challenge of understanding real-world documents organized in non-plain text formats (e.g., HTML and PDF).


\subsubsection{Visual instruction tuning.}
Researchers have recently explored the application of LLMs to vision-language tasks by distilling the output of LLMs~\cite{liu2023llava,zhu2023minigpt,ye2023mplugowl} or training with handcrafted instructions~\cite{xu-etal-2023-multiinstruct,instructblip}. However, as pointed out in~\cite{liu2023hidden}, these models struggle with tasks requiring document understanding abilities because they do not assume that text might be contained in images during instruction tuning. To mitigate this issue, LLaVAR~\cite{zhang2023llavar} and LLMDoc~\cite{ye2023mplugdocowl} fine-tune MLLMs with instruction tuning on document images. However, these approaches have trouble understanding diverse real-world documents because (i) the datasets provide a few document and task types, hindering zero-shot generalization; and (ii) the models simply encode documents via vision encoders and cannot explicitly learn document meta-information (e.g., document layout). In contrast, the InstructDoc covers diverse VDU tasks and open document types/formats, and InstructDr learns rich representations of the underlying structure of documents with instructions.

\section{InstructDoc Dataset}

\begin{figure*}[t!]
    \centering
\includegraphics[width=.9\textwidth]{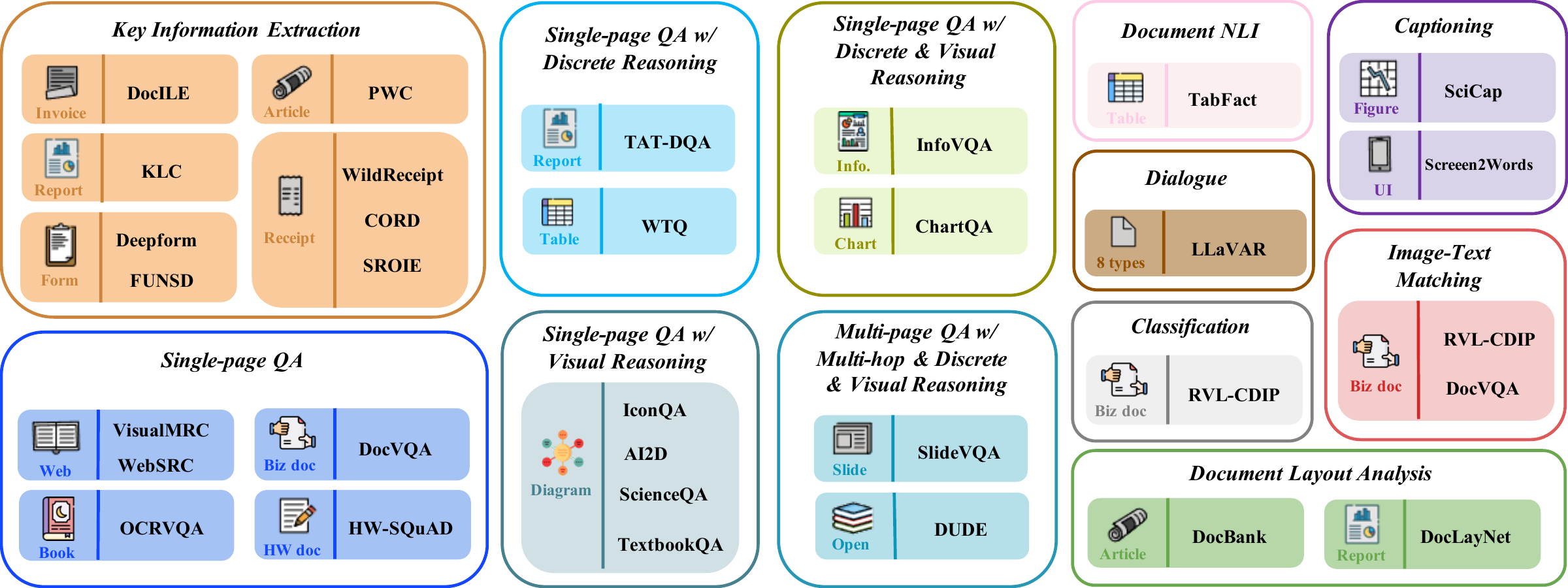}
    \caption{Datasets used in InstructDoc. InstructDoc covers a wide range of VDU tasks and document types and formats.}
    \label{fig:dataset}
\end{figure*}

\subsection{Problem Formulation}
All of the tasks in InstructDoc are simply defined as: 
given an instruction $T$ and a document image $I$, a model outputs an answer $A$. Each task is composed of one or more datasets, where the dataset $\mathcal{D}$ is associated with the set of $K$ instructions $\mathcal{T^{\mathcal{D}}} = \{T^{\mathcal{D}}_1, ..., T^{\mathcal{D}}_K\}$ and contains $N$ instances $\{(\mathcal{T^{\mathcal{D}}}, I_j, A_j)\}^{N}_{j=1}$. Here, we randomly select the instruction from $\mathcal{T^{\mathcal{D}}}$ for every instance. Note that we allow the utilization of external OCR engines to derive the answer in our setting, as in the previous VDU benchmark~\cite{borchmann2021due}. Our goal is to enable the model to perform a wide range of VDU tasks with instructions rather than improving the accuracy of text recognition~\cite{zhang2023llavar}.

We mainly evaluate the models' ability to perform zero-shot learning scenarios. Specifically, we fine-tune a model on a collection of instruction tasks and evaluate it on unseen datasets defined three types: (i) \textbf{Test$_{\text{Cross-Dataset}}$}: datasets not used during training, but whose tasks exist in training set; (ii) \textbf{Test$_{\text{Cross-Task}}$}: datasets and associated tasks entirely unseen during training; and (iii) \textbf{Test$_{\text{Cross-Domain}}$}: datasets, tasks, and document types entirely unseen during training.

\subsection{Dataset Collection}
In this section, we describe the collection process of the InstructDoc dataset. InstructDoc is designed to cover a wide range of VDU tasks with instructions that require reasoning among document layout, images, and text.

\subsubsection{Source dataset collection.}
Figure \ref{fig:dataset} shows the source datasets in InstructDoc. We collected 30 publicly available datasets and 12 tasks in VDU areas from DUE~\cite{borchmann2021due} as well as through manual searches. Following the task clusters defined in previous works~\cite{wei2021finetuned,instructblip}, we divided the QA datasets that require different reasoning abilities into different tasks. As a result, we divided the collected datasets into the following tasks:
\begin{itemize}
    \item \textbf{Key Information Extraction (KIE)} assigns each word a semantic entity label from  predefined categories~\cite{simsa2023docile,jaume2019funsd,sun2021spatial,park2019cord,huang2019icdar2019}.
    \item \textbf{Single-page QA} is a task of QA on single-page documents and focuses on document layout and textual content understanding~\cite{DBLP:conf/aaai/TanakaNY21,ChenZCJZLX021,MishraSSC19,Mathew_2021_WACV,tuselmann2022recognition}.
    \item \textbf{Single-page QA w/ Discrete Reasoning} requires various arithmetic abilities, including addition, sorting, or counting~\cite{zhu2022towards}. 
    \item \textbf{Single-page QA w/ Visual Reasoning} requires a set of abilities, including object (e.g., icon) recognition, commonsense understanding, and relation extraction on single-page documents~\cite{lu2021iconqa,kembhavi2016diagram,lu2022learn,kembhavi2016diagram}. 
    \item \textbf{Single-page QA w/ Discrete \& Visual Reasoning} requires both discrete and visual reasoning~\cite{Mathew_2022_WACV,masry-etal-2022-chartqa} on single-page documents.
    \item \textbf{Multi-page QA w/ Multi-hop \& Discrete \& Visual Reasoning} requires understanding the content relationship via multi-hop reasoning as well as discrete/visual reasoning on multi-page documents~\cite{SlideVQA2023,landeghem2023document}.
    \item \textbf{Document NLI} is a task of natural language inference that predicts the entailment relationship between two sentences in a document~\cite{borchmann2021due}
    \item \textbf{Dialogue} involves a human-agent interaction on the basis of document images~\cite{zhang2023llavar}.
    \item \textbf{Captioning} involves producing descriptions of documents~\cite{hsu-etal-2021-scicap-generating,wang2021screen2words}.
    \item \textbf{Classification} involves classifying a document from a set of candidate labels~\cite{harley2015evaluation}.
    \item \textbf{Document Layout Analysis (DLA)} determines a document’s components with bounding boxes~\cite{li-etal-2020-docbank,doclaynet}
    \item \textbf{Image-Text Matching (ITM)} requires the model to determine whether a given OCR text and image match. 
\end{itemize}

\begin{figure*}[t!]
    \centering
    \includegraphics[width=.82\textwidth]{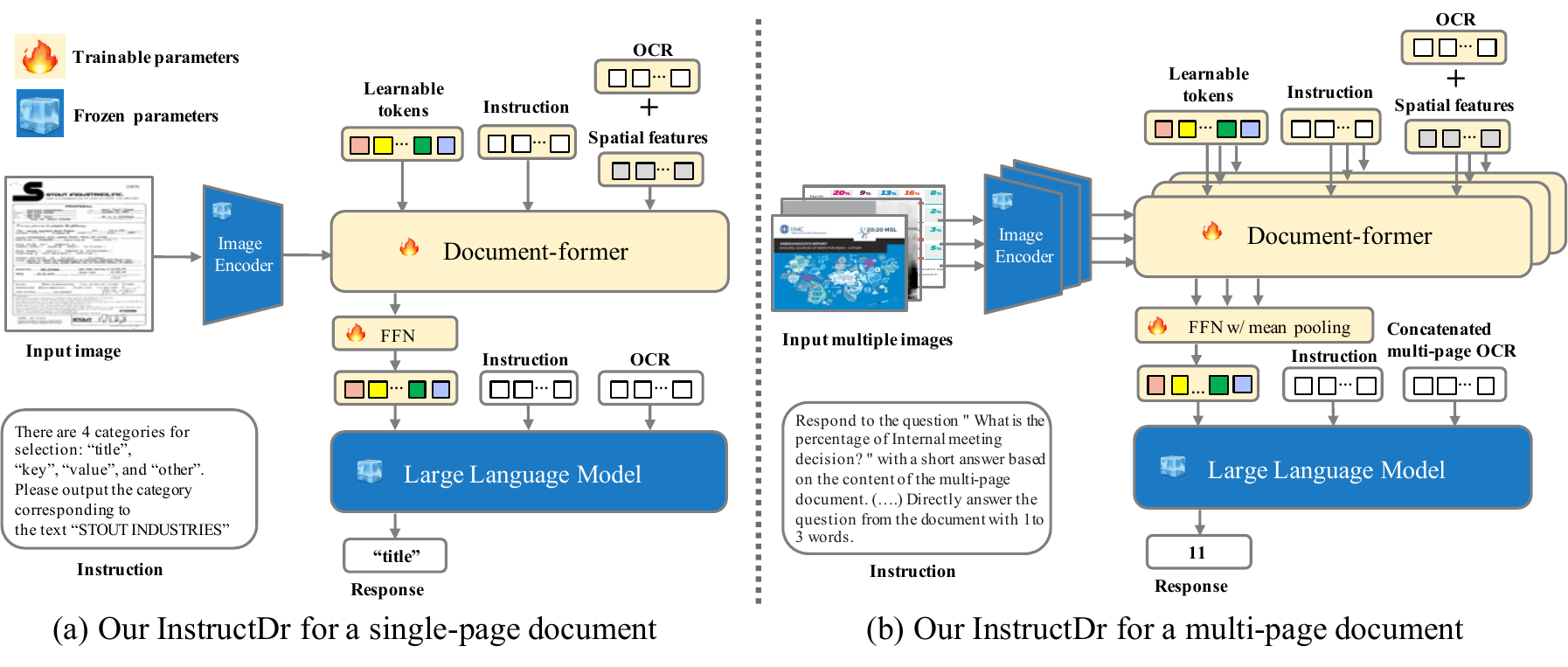}
    \caption{InstructDr model. We update only the parameters of Document-former and the projection FFN layer during training.}
    \label{fig:instructdlip}
\end{figure*}

\subsubsection{Query rephrasing.}
We found that two KIE datasets (FUNSD and CORD) are challenging because they contain abbreviated queries that are difficult for humans to comprehend. To bridge the gap between humans and machines, we replace these queries with complete and more easily understandable phrases (e.g., \texttt{menu.vatyn} $\to$ \texttt{menu\_whether\_price\_tax\_included}). 

\subsubsection{Instruction annotation.}
For each dataset, we manually crafted five to ten distinct instruction templates in a unified format. For QA tasks, the answers have diverse styles in the original datasets; 
for example, DocVQA's answer is extractive, which requires the model to extract a contiguous span of words from the document, but VisualMRC's answer is generative, which is not limited to the word spans. Hence, an instruction that sufficiently describes an arbitrary VDU task should include {\textit{intent}} and {\textit{answer style}} or only {\textit{intent}}. Specifically, as shown in Figure~\ref{fig:samples}, {\textit{intent}} describes how the task can be performed and {\textit{answer style}} describes how the model generates the output. If each dataset provides {\textit{query and options}}, we fill it in annotated instruction templates.

\subsubsection{Data split.} 
We split InstructDoc into 23 held-in and seven held-out datasets. For the held-out evaluation, we aim to understand how instruction tuning on the held-in datasets improves the zero-shot generalization performance on unseen datasets, including (i) \textbf{Test$_{\text{Cross-Dataset}}$}: FUNSD and CORD datasets, (ii) \textbf{Test$_{\text{Cross-Task}}$}: ChartQA, InfoVQA, and TabFact datasets, and (iii) \textbf{Test$_{\text{Cross-Domain}}$}: DUDE and SlideVQA datasets. All other datasets were held-in ones to train our model. Note that the held-out datasets were carefully selected in order to avoid data contamination.

\begin{table}[t!]
    \centering
        \scalebox{1.0}{
    \tabcolsep=1pt
    \small
    \begin{tabular}{lccccccc} 
        \toprule
         & LLaVAR & DocOwl & InstructDoc \\ \midrule
        Both Single/Multi-page docs & & & \checkmark \\ 
        Instruction annotation & & \checkmark & \checkmark\\
        Answer style annotation & & & \checkmark  \\
       \#Document types & 8 & 7 & Open \\ 
        \#Seed datasets & 1 & 8 & 30 \\
        \#Task clusters & 1 & 3 & 12 \\
        \#Avg.$_{\pm \text{Std.}}$ IT words & - & 5$_{\pm 0}$ & 20.3$_{\pm 11.2}$ \\
        \#Avg.$_{\pm \text{Std.}}$ IT & - & 1$_{\pm 0}$ & 7.4$_{\pm 2.4}$ \\
        \#Avg.$_{\pm \text{Std.}}$ OCR words & 52.5$_{\pm 93.1}$ & 270.1$_{\pm 807.2}$ & 443.2$_{\pm 1442.8}$ \\
        \#Avg.$_{\pm \text{Std.}}$ Answer words & 34.5$_{\pm 27.5}$ & 1.9$_{\pm  2.7}$ & 5.88$_{\pm 17.7}$ \\
        \bottomrule
    \end{tabular}
    }
    \caption{Statistics of InstructDoc and other VDU instruction tuning datasets. We excluded data other than the VDU tasks from DocOwl. IT denotes instruction templates.}
    \label{tab:comparison}
\end{table}

\subsection{Comparison with Related Datasets}
Table~\ref{tab:comparison} shows the statistics of InstructDoc and other VDU instruction tuning datasets, including LLaVAR~\cite{zhang2023llavar} and DocOwl~\cite{ye2023mplugdocowl}. InstructDoc has three unique key properties; First, it is the first dataset to address open document types, including multi-page documents and has the highest standard deviation in the number of OCR tokens (1442.8) compared with LLaVAR (93.1) and DocOwl (807.2). This implies that our dataset is a more challenging setting. Second, InstructDoc covers the widest range of tasks, offering four times more tasks compared with DocOwl, while LLaVAR provides only a single task. Finally, InstructDoc provides a more extensive set of instructions (20.3 words and 7.4 templates) and annotates various answer styles within the instructions to deal with various VDU tasks that require diverse abilities. In contrast, the instructions in DocOwl are limited (five words and a single template) and LLaVAR has only machine-generated instructions, and they may not generalize well to reformulations and new tasks.

\section{Our Model}

Figure \ref{fig:instructdlip} depicts our model, InstructDr
(\textbf{Instruct}ion-based \textbf{D}ument \textbf{r}eading and understanding model).
We use pre-trained BLIP-2~\cite{li2023blip2}, a state-of-the-art MLLM connected with instruction-tuned FlanT5~\cite{chung2022scaling}, as the base model. We extend BLIP-2 in three key ways; (i) equipping it with Document-former, an enhanced Q-former module that can capture and convert the visual and textual content/layout of documents into representations of the LLM, (ii) conducting multi-task instruction tuning with unified formats, and (iii) encoding multiple images in parallel to facilitate understanding of multi-page documents.

\begin{table*}[t!]
    \centering
        \scalebox{1.0}{
    \tabcolsep=2.8pt
    \small
    \begin{tabular}{lccccccccccc} 
        \toprule
        & & & & \multicolumn{2}{c}{Cross-Dataset} & \multicolumn{3}{c}{Cross-Task} & \multicolumn{2}{c}{Cross-Domain} \\ \cmidrule(lr){5-6} \cmidrule(lr){7-9} \cmidrule(lr){10-11}
        \multirow{2}{*}{Model} & \multirow{2}{*}{Modal} & \multirow{2}{*}{\#TuP} & \multirow{2}{*}{\#ToP} & FUNSD & CORD & ChartQA & InfoVQA & TabFact & DUDE & SlideVQA & Held-out \\ 
        & &  &  &  eF1/F1 & eF1/F1 &  RAcc./F1 & ANLS/F1 &  Acc./F1 & ANLS/F1 & EM/F1 & Avg.\\ \midrule
        LLMDoc & V & 388M & 7B & -/- &  -/- &  -/-& 38.2\dag/- & 60.2\dag/- & -/- & -/- & -/- \\ 
        LLaVA & TV & 13B & 13B & 12.0/1.3 & 0.2/ 5.1 & 0.0/1.7 & 3.4/3.5 &  0.0/0.0 & 6.5/5.9 & 0.0/2.3 & 3.1/2.8
        \\
        LLaVAR & TV & 13B & 13B & 12.0/2.0 & 0.1/10.8 & 0.0/3.0  & 6.2/4.6 & 0.0/2.1 & 8.1/5.1 & 0.0/6.2 & 3.8/4.8
        \\
        MiniGPT-4 & TV & 3.1M & 7B & 12.0/2.2 & 0.2/ 2.1 & 0.0/0.4 & 4.3/0.5 &  0.3/0.2 &  5.9/1.1 & 0.0/0.4 & 3.2/1.0 \\
        mPLUG-Owl & TV & 388M & 7B & 12.0/6.7 & 0.2/15.0 &  0.0/0.3 & 5.6/5.3 &  0.0/2.6 & 5.8/5.5 & 0.0/0.4 & 3.4/5.1 \\
        InstructBLIP & TV & 103M & 3.4B & 16.8/15.0 & 4.9/9.5 &  3.3/7.2 & 8.7/7.3 &  33.6/33.7 & 11.0/8.8 & 5.2/9.0 & 11.9/12.9 \\ 
        BLIP-2 & TV & 103M & 3.4B & 19.6/19.6 & 32.0/51.9 &  23.6/21.5 & 48.2/36.7 &  58.6/58.6 & 39.8/35.4 & 28.3/38.8 & 35.7/37.5 \\
        \midrule
        BLIP-2 trained on IDoc & TV & 103M & 3.4B & 26.0/26.1 & 33.8/54.7 &  24.7/21.2 & 47.8/35.4 &  58.8/58.8 & 43.9/40.4 & 30.1/38.8 & 37.9/39.3 \\
        InstructDr (Ours) & TLV & 103.1M & 3.4B & \textbf{38.2}/\textbf{38.1} & \textbf{46.0}/\textbf{62.7} & \textbf{29.4}/\textbf{22.3} & \textbf{50.9}/\textbf{37.6} &  \textbf{59.4}/\textbf{59.4} & \textbf{45.2}/\textbf{41.6} & \textbf{31.9}/\textbf{40.2} & \textbf{43.0}/\textbf{43.1} \\
        \bottomrule
    \end{tabular}
    }
    \caption{Zero-shot performance of InstructDr and MLLMs on VDU tasks. ``T/L/V" denotes the ``text/layout/visual" modality of documents. \#TuP/\#ToP denotes the number of tuning/total parameters. The highest zero-shot performances are marked in bold. \dag denotes the supervised performance reported in the original paper, as it is not publicly available. IDoc denotes InstructDoc.}
    \label{tab:main}
\end{table*}

\subsection{Spatial-aware Document Feature Extraction}
\subsubsection{Document image/OCR and instruction encoding.}
To encode a document image, we use a pre-trained CLIP~\cite{radford2021learning} vision encoder to extract its visual features $\mathbf{z}^{\text{vis}}$. Additionally, we process the document image using an OCR engine and apply a sub-word tokenizer to obtain $M$ word tokens $\{s_i\}_{i=1}^M$ and their corresponding bounding boxes $\{ (x_i^1, y_i^1, x_i^2, y_i^2)\}_{i=1}^M$, where ($x^1$, $y^1$) and ($x^2$, $y^2$) represent the coordinates of the top-left and bottom-right corners, respectively. To learn the visual layout of the image, we construct a spatially aware OCR representation $\mathbf{z}_i^{\text{ocr}} = \mathbf{z}_i^{\text{word}} + \mathbf{z}_i^{\text{bbox}}$ with learnable embedding layers $\mathbf{W}^{\{s, x, y, h, w\}}$, where OCR text features are calculated as $\mathbf{z}^{\text{word}}_i = \mathbf{W}^s(s_i)$ and spatial features are calculated as $\mathbf{z}^{\text{bbox}}_i = \mathbf{W}^x(x^1_i, x^2_i) + \mathbf{W}^y(y^1_i, y^2_i) + \mathbf{W}^h(y^2_i - y^1_i) + \mathbf{W}^w(x^2_i - x^1_i)$. Similarly, we encode an instruction by $\mathbf{W}^{s}$ and obtain its features $\mathbf{z}^{\text{ins}}$.
 
\subsubsection{Document-former.}
We introduce Document-former, which is a trainable module to bridge the gap between an image encoder and an LLM, enabling extraction of document content/layout that LLMs can understand. The architecture of Document-former is a stack of Transformer blocks with cross-attention layers. To map document features into the LLM's space, we use a set of $m$ learnable tokens $\mathbf{z}^{\text{token}} \in \mathbb{R}^{d}$, where $d$ is the dimension of the hidden size. These tokens $\mathbf{z}^{\text{token}}$ interact with $\mathbf{z}^{\text{vis}}$ through cross-attention layers and with the input sequence, composed of $\mathbf{z}^{\text{ins}}$ and $\mathbf{z}^{\text{ocr}}$, through self-attention layers. As a result, we obtain $\mathbf{z}^{\text{doc}}$ and transform it via a projection feed-forward network (FFN) layer to $\mathbf{h}^{\text{doc}} \in \mathbb{R}^{m \times d^{\text{LLM}}}$, which have the same dimension $d^{\text{LLM}}$ as the LLM’s input embedding.


\subsection{Multimodal Document Large Language Model} 

\subsubsection{Connecting document features to LLM.}
The LLM receives the document embeddings $\mathbf{h}^{\text{doc}}$, the instruction, and OCR tokens as input and outputs the answer $\mathbf{A}$, token by token. 
The parameters of the LLM are initialized from an instruction-tuned FlanT5.

\subsubsection{Parameter-efficient multi-task instruction tuning.}
To achieve task-agnostic learning, we formulate the process of learning all held-in tasks in a unified sequence-to-sequence abstraction through instructions. To train the LLM efficiently, we update only the parameters of the Document-former (including $\mathbf{W}^{\{s, x, y, h, w\}}$) and the projection FFN layer, while keeping other parameters frozen during training. We optimize the model by minimizing the negative log-likelihood between the ground-truth and predictions.

\subsubsection{Multi-page document understanding.} We also support performing reasoning across multiple document pages. As shown in Figure~\ref{fig:instructdlip}b, each image is processed individually by the image encoder and Document-former, and their resulting document embeddings are mean-pooled together before being fed into the LLM. The OCR input to the LLM consists of concatenated tokens extracted from each page.
\section{Experiments}

\begin{table*}[t!]
    \centering
        \scalebox{1.0}{
    \tabcolsep=3pt
    \small
    \begin{tabular}{lccccccccc} 
        \toprule
        &  & \multicolumn{2}{c}{Cross-Dataset} & \multicolumn{3}{c}{Cross-Task} & \multicolumn{2}{c}{Cross-Domain} \\ \cmidrule(lr){3-4} \cmidrule(lr){5-7} \cmidrule(lr){8-9}
        \multirow{2}{*}{Model} & \multirow{2}{*}{Modal} & FUNSD & CORD & ChartQA & InfoVQA & TabFact & DUDE & SlideVQA & Held-out \\ 
        &  &  eF1/F1 & eF1/F1 &  RAcc./F1 & ANLS/F1 &  Acc./F1 & ANLS/F1 & EM/F1 & Avg.\\ \midrule
        Supervised SOTA models & TLV & 92.1/- & 97.7/-  & 72.3/- & 54.8*/- &  83.2*/- & 46.1*/- & 33.5/41.7 & -/-  \\ \midrule
        ChatGPT & T& 21.8/21.2 & 30.4/49.3
        &  16.0/16.8 & 37.8/29.5 &  52.5/52.4 & 34.5/32.3 & 11.7/23.8 & 29.2/32.2  \\
        GPT-4 & T& 47.5/47.5 & 69.4/81.7
        &  20.9/27.6 & 49.9/46.5 &  68.8/68.8 & 46.3/45.1 & 21.0/36.4 &46.3/50.5  \\ \midrule 
        InstructDr (Ours) & TLV & 38.2/38.1 & 46.0/62.7 & 29.4/22.3 & 50.9/37.6 &  59.4/59.4 & 45.2/                                        41.6 & 31.9/40.2 & 43.0/43.1 \\
        \bottomrule
    \end{tabular}
    }
    \caption{Zero-shot performance on VDU tasks of InstructDr and supervised SOTA models and powerful text-based LLMs. * denotes the performance on different splits we used because they evaluated on the leaderboard and F1 cannot be used.}
    \label{tab:compare_chatgpt}
\end{table*}

\subsection{Experimental Setup}
We mainly conducted evaluations under three zero-shot settings, including \textbf{Test$_{\text{Cross-Dataset}}$}, \textbf{Test$_{\text{Cross-Task}}$}, and \textbf{Test$_{\text{Cross-Domain}}$}. Furthermore, we evaluated our model under the task-specific fine-tuning setting.

\subsubsection{Baselines.}
We compared InstructDr with seven 
state-of-the-art (SOTA) MLLMs, including \textbf{LLaVA}~\cite{liu2023llava}, \textbf{MiniGPT-4}~\cite{zhu2023minigpt} and \textbf{mPLUG-Owl}~\cite{ye2023mplugowl}, which align CLIP visual encoder with Vicuna~\cite{vicuna2023} trained on 
a dialogue generated by GPT-4~\cite{openai2023gpt4}; \textbf{BLIP-2}~\cite{li2023blip2}, which connects a 
FlanT5 with a vision encoder; \textbf{InstructBLIP}~\cite{instructblip}, which fine-tunes BLIP-2 with instructions on scene images; and \textbf{LLMDoc}~\cite{ye2023mplugdocowl} and \textbf{LLaVAR}~\cite{zhang2023llavar}, which fine-tune mPULG-Owl/LLaVA on the DocOwl/LLaVAR datasets. Additionally, we used \textbf{Supervised SOTA models}~\cite{appalaraju2023docformerv2,chen2023pali,huang2022layoutlmv3,landeghem2023document} on each dataset and two text-based LLMs, \textbf{ChatGPT} (\texttt{gpt-3.5-turbo-0613}) and \textbf{GPT-4}. To control the answer's length, we added control phrases (e.g., \textit{use 1 to 3 words to answer}) to the selected instructions. 

\subsubsection{Evaluation metrics.}
We followed the evaluation protocol of each dataset, we used \textbf{ANLS}~\cite{BitenTMBRJVK19} for InfoVQA, DUDE, Text-VQA and ST-VQA, \textbf{EM} for SlideVQA, Relaxed Accuracy (\textbf{RAcc.}) for ChartQA, entity F1 (\textbf{eF1}) for FUNSD and CORD, Accuracy (\textbf{Acc.}) for TabFact, and \textbf{ROUGE-L} for VisualMRC as evaluation metrics. Additionally, we used \textbf{F1} as the optional metrics.

\begin{table}[t!]
    \centering
        \scalebox{1.0}{
    \tabcolsep=2.3pt
    \small
    \begin{tabular}{lSSSS} 
        \toprule
        & {CORD} & {TabFact} & {DUDE} & {Held-out} \\ 
        Model & {eF1} & {Acc.} & {ANLS} & {Avg.}  \\ 
        \midrule
        InstructDr & \hspace{-0.15cm}\textbf{46.0} & \hspace{-0.15cm}\textbf{59.4}& \hspace{-0.15cm}\textbf{45.2} &  \hspace{-0.15cm}\textbf{43.0}  \\ \midrule
        w/o Document-former & 38.5 & 58.8 & 44.6 &  40.2 \\ 
        w/o Spatially OCR features & 33.8 & 58.8 & 43.9 &  37.9  \\ 
        w/o Mean pooling (concat.) & {-} & {-} &43.8 & {-} \\ \midrule
        w/o Instructions in test & 24.0 & 4.0 & 38.9 &  28.0 \\
        w/o Instructions in train & 17.3 & 58.2 & 34.0 &  28.9\\
        w/o Instructions in both & 0.4  & 3.7 & 24.4 &  21.3 \\
        w/o Query rephrasing & 30.9 & {-} & {-} &   {-} \\
        w/o Answer style annotation & {-} & {-} & 44.2 & {-} \\
        \bottomrule
    \end{tabular}
    }
    \caption{Ablation study of the architecture and instructions. We report the scores when the ablation can be conducted.}
    \label{tab:ablation}
\end{table}
\subsubsection{Implementation details.}
Following~\cite{wei2021finetuned}, we balanced the training instances of different tasks by sampling a maximum of 5k instances for each held-in dataset while keeping all evaluation instances. We used the AdamW~\cite{loshchilov2017decoupled} with 
a weight decay of 0.05. We applied a linear warmup during the initial 1,000 steps and used a cosine learning rate decay with a minimum learning rate of 0. We set the number of learnable tokens $m$ to $32$. 
All images of the model input were resized to $224$. We trained on eight A100 (40G) GPUs for three epochs and completed the training within two hours. If each dataset does not provide OCR, we extracted it via the Google Vision API.

\subsection{Experimental Results and Analysis}
\subsubsection{Does our model outperform existing MLLMs?}
Table~\ref{tab:main} shows that our model achieved the highest performance on all datasets compared with other MLLMs. InstructDr consistently outperformed its original backbone, BLIP-2, by a significant margin, indicating that instruction tuning on InstructDoc effectively enhances performance on unseen VDU datasets, tasks, and domains. In contrast, InstructBLIP, which is instruction-tuned BLIP-2 trained on scene images, performed worse than BLIP-2. This is because that InstructBLIP does not assume that the images might contain text during instruction tuning. BLIP-2 fine-tuned on InstructDoc falls short of achieving the same level of performance compared with InstructDr, indicating that InstructDr is better suited for comprehending diverse real-world documents. This conclusion is further supported by the results presented in Table~\ref{tab:ablation}, where ablations of Document-former, spatial information, and strategy of gathering multi-page features have a significant negative impact on performance.

\subsubsection{How well does our model perform in comparison with supervised SOTA models and powerful LLMs?}
As shown in Table~\ref{tab:compare_chatgpt}, our model outperformed ChatGPT on all datasets. Additionally, InstructDr achieved competitive results with supervised SOTA models and GPT-4 on the DUDE and SlideVQA datasets that require multiple reasoning skills (e.g., discrete, visual, and multi-hop reasoning). This indicates that our model can effectively learn diverse skills through instruction tuning with InstructDoc.

\begin{figure}[t!]
    \centering
    \includegraphics[width=.44\textwidth]{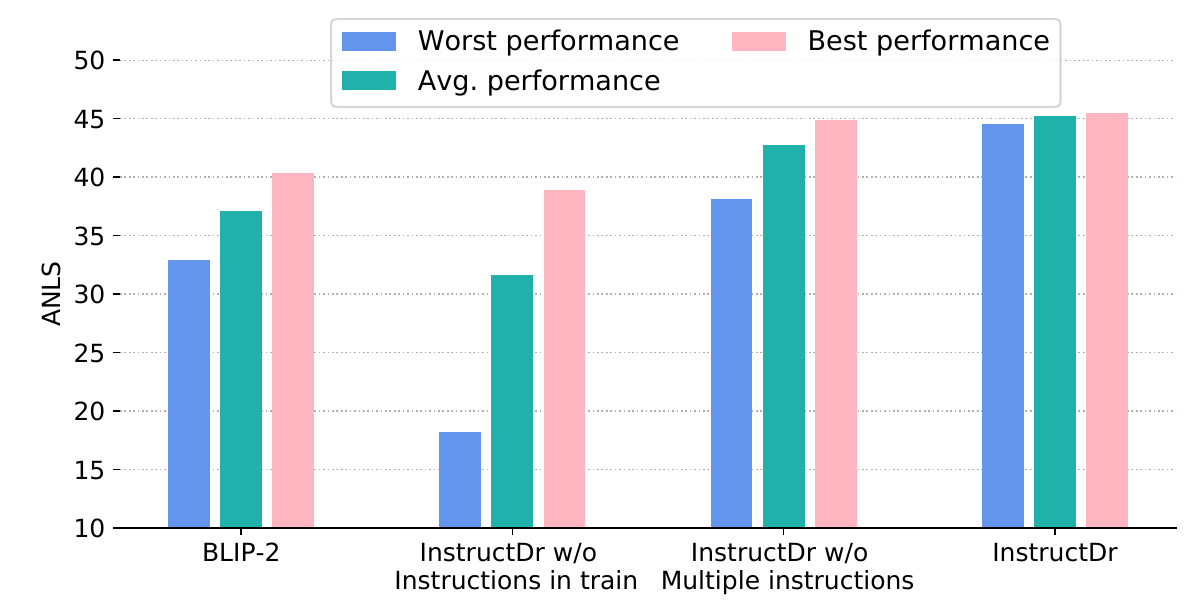}
    \caption{Comparison of zero-shot performance on DUDE for five different instructions.  w/o Multiple instructions denotes our model trained with a single instruction per dataset.}
    \label{fig:robustnss}
\end{figure}

\begin{figure}[t!]
    \centering
    \includegraphics[width=.44\textwidth]{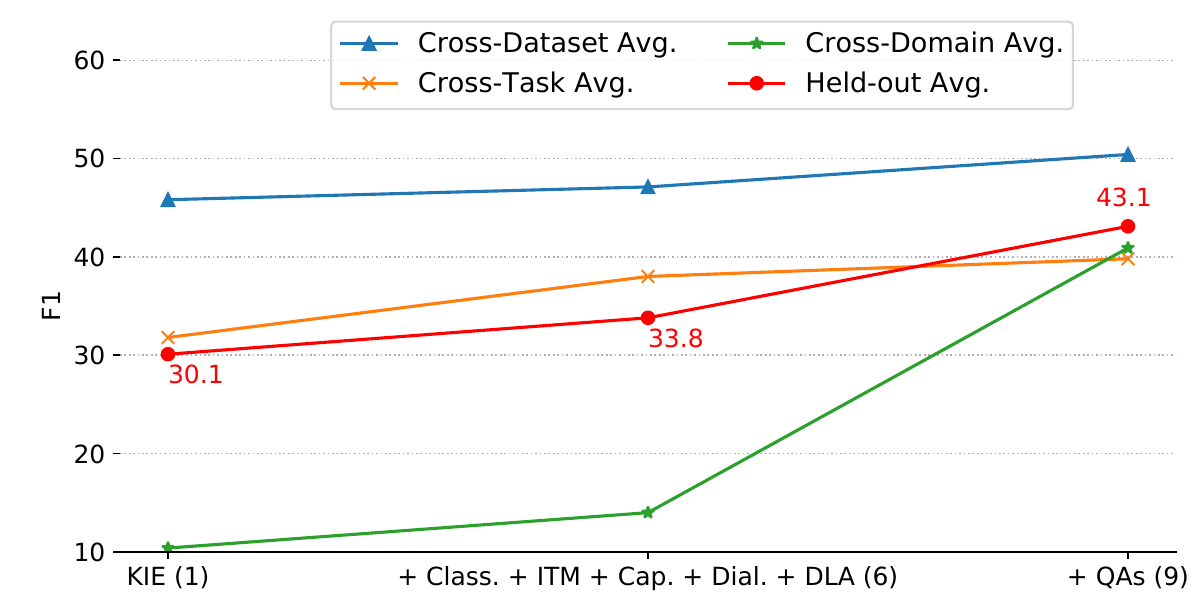}
    \caption{Model performance as the number of task clusters used in training. ($\cdot$) denotes the
number of tasks.}
    \label{fig:ablation_task}
\end{figure}

\subsubsection{What is the role of instructions?}
As shown in Table~\ref{tab:ablation}, removing instructions (i.e., only \textit{query and options} as the model input)
significantly decreased zero-shot performance during training or/and test time, indicating the effectiveness of incorporating instructions. This result was observed on the high-quality instruction-tuning datasets~\cite{wei2021finetuned,xu-etal-2023-multiinstruct}. Moreover, our instruction annotations, including query rephrasing and answer styles, helped to improve the zero-shot performance. 

\begin{figure*}[t!]
    \centering
    \includegraphics[width=.93\textwidth]{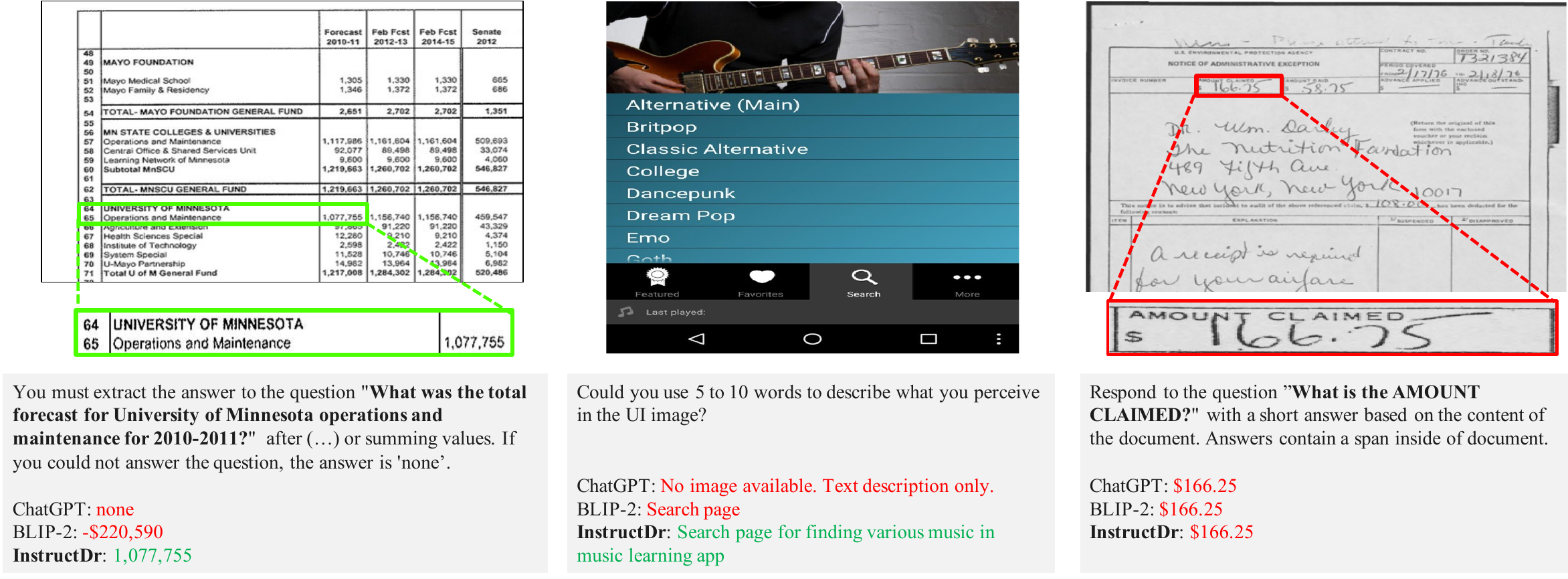}
    \caption{Qualitative examples. Outputs are \textcolor{mygreen}{correct/sufficient} and \textcolor{red}{incorrect/insufficient} answers. (...) denotes ellipsis.}
    \label{fig:output}
\end{figure*}

\subsubsection{Does our model have robustness towards diverse instructions?}
Figure~\ref{fig:robustnss} shows the performance variance when the models were given five different instructions; InstructDr exhibited the smallest performance variance and outperformed the other models. This indicates InstructDoc empowers the model with the ability to deal with a variety of instructions. Our results also suggest that using multiple instructions per dataset is important for achieving decent performance. 

\subsubsection{What is the impact of diverse task clusters?}
As shown in Figure~\ref{fig:ablation_task}, as the number of task clusters increases, we can observe an improvement in models’ zero-shot performance.

\begin{table}[t!]
    \centering
        \scalebox{1.0}{
    \tabcolsep=3pt
    \small
    \begin{tabular}{lccccc} 
        \toprule
         & VisualMRC & DUDE & \multicolumn{2}{c}{SlideVQA}  \\ 
        Model & ROUGE-L & ANLS & EM & F1 \\   \midrule
        Supervised SOTA models & 52.2 & 46.1 & 33.5 & 41.7  \\
        \midrule
        BLIP-2 & 60.5 & 45.6 & 36.9 & 46.5 \\
        InstructDr & \textbf{61.1} & \textbf{46.8} & \textbf{37.7} & \textbf{47.3}  \\ 
        \bottomrule
    \end{tabular}
    }
    \caption{Fine-tuning performance in held-in (VisualMRC) and held-out (DUDE, SlideVQA) tasks on the test set.}
    \label{tab:finetune}
\end{table}

\begin{table}[t!]
    \centering
        \scalebox{1.0}{
    \tabcolsep=3pt
    \small
    \begin{tabular}{lcccc} 
        \toprule
        & Image type in & \multicolumn{2}{c}{TextVQA} & ST-VQA  \\ 
        Model & instruction tuning & Acc. & ANLS & ANLS \\ 
        \midrule
        BLIP-2 & - & 48.7 & 64.8 & 39.1 \\
        InstructBLIP & Daily scene & 52.8 & 67.3 & \textbf{45.7} \\ \midrule
        InstructDr & Documents & \textbf{53.8} & \textbf{68.1} & 43.3 \\ 
        \bottomrule
    \end{tabular}
    }
    \caption{Zero-shot performance of scene-text VQA task.}
    \label{tab:textvqa}
\end{table}

\subsubsection{Are our model weights effective for task-specific fine-tuning?}
We further fine-tuned InstructDr (only Document-former module) on a specific dataset to investigate the knowledge and transferability of our instruction-tuned model weights. Table~\ref{tab:finetune} shows the fine-tuning performance on held-in (VisualMRC) and held-out (DUDE, SlideVQA) tasks. InstructDr achieved state-of-the-art finetuning performance on VisualMRC, DUDE, and SlideVQA using a unified model. Compared with BLIP-2, InstructDr exhibited superior fine-tuning performance on both held-in/out datasets, validating InstructDr as a better weight initialization model for task-specific fine-tuning.

\subsubsection{Can our model also understand images other than documents?}
Table~\ref{tab:textvqa} shows the zero-shot performance of scene-text VQA~\cite{SinghNSJCBPR19,BitenTMBRJVK19} on scene images, which are the unseen image types in InstructDoc but were used for training our base model, BLIP-2. Note that ST-VQA's images include the part of COCO~\cite{lin2014microsoft} that InstructBLIP was trained on. This result indicates that InstructDr can effectively learn visual reasoning skills without forgetting the abilities of the original backbone.

\subsubsection{Qualitative examples.}
Figure~\ref{fig:output} visualizes output examples, where the left/center/right examples require table/visual/hand-written text understanding skills. ChatGPT gave incorrect answers because it can only consider text information. Moreover, while BLIP-2 could not follow instructions (e.g., \textit{use 5 to 10 words}) and extract items from structured text, InstructDr accomplished diverse VDU tasks with instructions. As shown in the right example, all models affected OCR quality, causing incorrect answers. 

\section{Limitations}
Despite its impressive performance on various VDU tasks with instructions, InstructDr suffers from noisy OCR predictions, whose performance depends highly on OCR text qualities (right of Figure~\ref{fig:output}). We argue that our approach is more cost-efficient and accurate because another approach, the pixel-based ones~\cite{kim2022ocr,chen2023pali}, requires a large amount of computation to encode high-resolution images and cannot use document meta-information (e.g., bounding boxes). Moreover, since InstructDoc only contains a single document-text pair per instance, it cannot learn the correlation among multiple document-text pairs and lacks an in-context learning capability. The same observation has also been reported in the Flamingo~\cite{alayrac2022flamingo} and BLIP-2. Finally, while we have constructed diverse VDU tasks, the number of tasks and corresponding instructions are still limited. We plan to consider utilizing automatic generation and augmentation techniques to increase the variety of instructions available.

\section{Conclusion}
We introduced a new large-scale instruction-tuning dataset, InstructDoc, to lay the foundation for building general-purpose VDU models that can follow natural language instructions. We also introduced a simple yet effective instruction tuning model, InstructDr, which unifies the vision, text, and layout modalities of documents by bridging the gap between a vision encoder and an LLM with Document-former. We performed a comprehensive study on instruction tuning with InstructDoc and demonstrated its generalization capability to a wide range of VDU datasets, tasks, and domains with instructions. 
We believe that our dataset will facilitate research on developing general-purpose document artificial intelligence systems.

\newpage
\bibliography{references}

\begin{thebibliography}{58}
\providecommand{\natexlab}[1]{#1}

\bibitem[{Alayrac et~al.(2022)Alayrac, Donahue, Luc, Miech, Barr, Hasson, Lenc, Mensch, Millican, Reynolds et~al.}]{alayrac2022flamingo}
Alayrac, J.-B.; Donahue, J.; Luc, P.; Miech, A.; Barr, I.; Hasson, Y.; Lenc, K.; Mensch, A.; Millican, K.; Reynolds, M.; et~al. 2022.
\newblock Flamingo: a Visual Language Model for Few-Shot Learning.
\newblock In \emph{NeurIPS}.

\bibitem[{Appalaraju et~al.(2021)Appalaraju, Jasani, Kota, Xie, and Manmatha}]{appalaraju2021docformer}
Appalaraju, S.; Jasani, B.; Kota, B.~U.; Xie, Y.; and Manmatha, R. 2021.
\newblock Docformer: End-to-End Transformer for Document Understanding.
\newblock In \emph{CVPR}, 993--1003.

\bibitem[{Appalaraju et~al.(2023)Appalaraju, Tang, Dong, Sankaran, Zhou, and Manmatha}]{appalaraju2023docformerv2}
Appalaraju, S.; Tang, P.; Dong, Q.; Sankaran, N.; Zhou, Y.; and Manmatha, R. 2023.
\newblock DocFormerv2: Local Features for Document Understanding.
\newblock \emph{arXiv:2306.01733}.

\bibitem[{Bach et~al.(2022)Bach, Sanh, Yong, Webson, Raffel, Nayak, Sharma, Kim, Bari, Fevry, Alyafeai, Dey, Santilli, Sun, Ben-david, Xu, Chhablani, Wang, Fries, Al-shaibani, Sharma, Thakker, Almubarak, Tang, Radev, Jiang, and Rush}]{bach-etal-2022-promptsource}
Bach, S.; Sanh, V.; Yong, Z.~X.; Webson, A.; Raffel, C.; Nayak, N.~V.; Sharma, A.; Kim, T.; Bari, M.~S.; Fevry, T.; Alyafeai, Z.; Dey, M.; Santilli, A.; Sun, Z.; Ben-david, S.; Xu, C.; Chhablani, G.; Wang, H.; Fries, J.; Al-shaibani, M.; Sharma, S.; Thakker, U.; Almubarak, K.; Tang, X.; Radev, D.; Jiang, M. T.-j.; and Rush, A. 2022.
\newblock PromptSource: An Integrated Development Environment and Repository for Natural Language Prompts.
\newblock In \emph{ACL-demo}, 93--104.

\bibitem[{Biten et~al.(2019)Biten, Tito, Mafla, i~Bigorda, Rusi{\~{n}}ol, Jawahar, Valveny, and Karatzas}]{BitenTMBRJVK19}
Biten, A.~F.; Tito, R.; Mafla, A.; i~Bigorda, L.~G.; Rusi{\~{n}}ol, M.; Jawahar, C.~V.; Valveny, E.; and Karatzas, D. 2019.
\newblock Scene Text Visual Question Answering.
\newblock In \emph{ICCV}, 4290--4300.

\bibitem[{Borchmann et~al.(2021)Borchmann, Pietruszka, Stanislawek, Jurkiewicz, Turski, Szyndler, and Grali{\'n}ski}]{borchmann2021due}
Borchmann, {\L}.; Pietruszka, M.; Stanislawek, T.; Jurkiewicz, D.; Turski, M.; Szyndler, K.; and Grali{\'n}ski, F. 2021.
\newblock DUE: End-to-End Document Understanding Benchmark.
\newblock In \emph{NeurIPS}.

\bibitem[{Chen et~al.(2023)Chen, Djolonga, Padlewski, Mustafa, Changpinyo, Wu, Ruiz, Goodman, Wang, Tay et~al.}]{chen2023pali}
Chen, X.; Djolonga, J.; Padlewski, P.; Mustafa, B.; Changpinyo, S.; Wu, J.; Ruiz, C.~R.; Goodman, S.; Wang, X.; Tay, Y.; et~al. 2023.
\newblock PaLI-X: On Scaling up a Multilingual Vision and Language Model.
\newblock \emph{arXiv:2305.18565}.

\bibitem[{Chen et~al.(2021)Chen, Zhao, Chen, Ji, Zhang, Luo, Xiong, and Yu}]{ChenZCJZLX021}
Chen, X.; Zhao, Z.; Chen, L.; Ji, J.; Zhang, D.; Luo, A.; Xiong, Y.; and Yu, K. 2021.
\newblock WebSRC: {A} Dataset for Web-Based Structural Reading Comprehension.
\newblock In \emph{EMNLP}, 4173--4185.

\bibitem[{Chiang et~al.(2023)Chiang, Li, Lin, Sheng, Wu, Zhang, Zheng, Zhuang, Zhuang, Gonzalez, Stoica, and Xing}]{vicuna2023}
Chiang, W.-L.; Li, Z.; Lin, Z.; Sheng, Y.; Wu, Z.; Zhang, H.; Zheng, L.; Zhuang, S.; Zhuang, Y.; Gonzalez, J.~E.; Stoica, I.; and Xing, E.~P. 2023.
\newblock Vicuna: An Open-Source Chatbot Impressing GPT-4 with 90\%* ChatGPT Quality.

\bibitem[{Chung et~al.(2022)Chung, Hou, Longpre, Zoph, Tay, Fedus, Li, Wang, Dehghani, Brahma et~al.}]{chung2022scaling}
Chung, H.~W.; Hou, L.; Longpre, S.; Zoph, B.; Tay, Y.; Fedus, W.; Li, E.; Wang, X.; Dehghani, M.; Brahma, S.; et~al. 2022.
\newblock Scaling Instruction-Finetuned Language Models.
\newblock \emph{arXiv:2210.11416}.

\bibitem[{Dai et~al.(2023)Dai, Li, Li, Tiong, Zhao, Wang, Li, Fung, and Hoi}]{instructblip}
Dai, W.; Li, J.; Li, D.; Tiong, A. M.~H.; Zhao, J.; Wang, W.; Li, B.; Fung, P.; and Hoi, S. 2023.
\newblock InstructBLIP: Towards General-purpose Vision-Language Models with Instruction Tuning.
\newblock \emph{arXiv:2305.06500}.

\bibitem[{Harley, Ufkes, and Derpanis(2015)}]{harley2015evaluation}
Harley, A.~W.; Ufkes, A.; and Derpanis, K.~G. 2015.
\newblock Evaluation of Deep Convolutional Nets for Document Image Classification and Retrieval.
\newblock In \emph{ICDAR}, 991--995.

\bibitem[{Hsu, Giles, and Huang(2021)}]{hsu-etal-2021-scicap-generating}
Hsu, T.-Y.; Giles, C.~L.; and Huang, T.-H. 2021.
\newblock {S}ci{C}ap: Generating Captions for Scientific Figures.
\newblock In \emph{EMNLP Findings}, 3258--3264.

\bibitem[{Huang et~al.(2022)Huang, Lv, Cui, Lu, and Wei}]{huang2022layoutlmv3}
Huang, Y.; Lv, T.; Cui, L.; Lu, Y.; and Wei, F. 2022.
\newblock LayoutLMv3: Pre-training for Document AI with Unified Text and Image Masking.
\newblock In \emph{ACMM}, 4083--4091.

\bibitem[{Huang et~al.(2019)Huang, Chen, He, Bai, Karatzas, Lu, and Jawahar}]{huang2019icdar2019}
Huang, Z.; Chen, K.; He, J.; Bai, X.; Karatzas, D.; Lu, S.; and Jawahar, C. 2019.
\newblock ICDAR2019 Competition on Scanned Receipt OCR and Information Extraction.
\newblock In \emph{ICDAR}, 1516--1520.

\bibitem[{Iyer et~al.(2022)Iyer, Lin, Pasunuru, Mihaylov, Simig, Yu, Shuster, Wang, Liu, Koura et~al.}]{iyer2022opt}
Iyer, S.; Lin, X.~V.; Pasunuru, R.; Mihaylov, T.; Simig, D.; Yu, P.; Shuster, K.; Wang, T.; Liu, Q.; Koura, P.~S.; et~al. 2022.
\newblock OPT-IML: Scaling Language Model Instruction Meta Learning through the Lens of Generalization.
\newblock \emph{arXiv:2212.12017}.

\bibitem[{Jaume, Ekenel, and Thiran(2019)}]{jaume2019funsd}
Jaume, G.; Ekenel, H.~K.; and Thiran, J.-P. 2019.
\newblock FUNSD: A Dataset for Form Understanding in Noisy Scanned Documents.
\newblock In \emph{ICDARW}.

\bibitem[{Kembhavi et~al.(2016)Kembhavi, Salvato, Kolve, Seo, Hajishirzi, and Farhadi}]{kembhavi2016diagram}
Kembhavi, A.; Salvato, M.; Kolve, E.; Seo, M.; Hajishirzi, H.; and Farhadi, A. 2016.
\newblock A Diagram Is Worth A Dozen Images.
\newblock In \emph{ECCV}, 235--251.

\bibitem[{Kembhavi et~al.(2017)Kembhavi, Seo, Schwenk, Choi, Farhadi, and Hajishirzi}]{KembhaviSSCFH17}
Kembhavi, A.; Seo, M.~J.; Schwenk, D.; Choi, J.; Farhadi, A.; and Hajishirzi, H. 2017.
\newblock Are You Smarter Than a Sixth Grader? Textbook Question Answering for Multimodal Machine Comprehension.
\newblock In \emph{CVPR}, 5376--5384.

\bibitem[{Kim et~al.(2022)Kim, Hong, Yim, Nam, Park, Yim, Hwang, Yun, Han, and Park}]{kim2022ocr}
Kim, G.; Hong, T.; Yim, M.; Nam, J.; Park, J.; Yim, J.; Hwang, W.; Yun, S.; Han, D.; and Park, S. 2022.
\newblock OCR-free Document Understanding Transformer.
\newblock In \emph{ECCV}, 498--517.

\bibitem[{Krishna et~al.(2017)Krishna, Zhu, Groth, Johnson, Hata, Kravitz, Chen, Kalantidis, Li, Shamma, Bernstein, and Fei{-}Fei}]{KrishnaZGJHKCKL17}
Krishna, R.; Zhu, Y.; Groth, O.; Johnson, J.; Hata, K.; Kravitz, J.; Chen, S.; Kalantidis, Y.; Li, L.; Shamma, D.~A.; Bernstein, M.~S.; and Fei{-}Fei, L. 2017.
\newblock Visual Genome: Connecting Language and Vision Using Crowdsourced Dense Image Annotations.
\newblock \emph{Int. J. Comput. Vis.}, 123(1): 32--73.

\bibitem[{Kuznetsova et~al.(2020)Kuznetsova, Rom, Alldrin, Uijlings, Krasin, Pont-Tuset, Kamali, Popov, Malloci, Kolesnikov et~al.}]{kuznetsova2020open}
Kuznetsova, A.; Rom, H.; Alldrin, N.; Uijlings, J.; Krasin, I.; Pont-Tuset, J.; Kamali, S.; Popov, S.; Malloci, M.; Kolesnikov, A.; et~al. 2020.
\newblock The Open Images Dataset V4: Unified image classification, object detection, and visual relationship detection at scale.
\newblock \emph{IJCV}, 1956--1981.

\bibitem[{Landeghem et~al.(2023)Landeghem, Tito, Borchmann, Pietruszka, J{\'o}ziak, Powalski, Jurkiewicz, Coustaty, Ackaert, Valveny et~al.}]{landeghem2023document}
Landeghem, J.; Tito, R.; Borchmann, {\L}.; Pietruszka, M.; J{\'o}ziak, P.; Powalski, R.; Jurkiewicz, D.; Coustaty, M.; Ackaert, B.; Valveny, E.; et~al. 2023.
\newblock Document Understanding Dataset and Evaluation (DUDE).
\newblock \emph{arXiv:2305.08455}.

\bibitem[{Lee et~al.(2023)Lee, Joshi, Turc, Hu, Liu, Eisenschlos, Khandelwal, Shaw, Chang, and Toutanova}]{lee2023pix2struct}
Lee, K.; Joshi, M.; Turc, I.~R.; Hu, H.; Liu, F.; Eisenschlos, J.~M.; Khandelwal, U.; Shaw, P.; Chang, M.-W.; and Toutanova, K. 2023.
\newblock Pix2Struct: Screenshot Parsing as Pretraining for Visual Language Understanding.
\newblock In \emph{ICML}, 18893--18912.

\bibitem[{Li et~al.(2023)Li, Li, Savarese, and Hoi}]{li2023blip2}
Li, J.; Li, D.; Savarese, S.; and Hoi, S. 2023.
\newblock {BLIP-2:} Bootstrapping Language-Image Pre-training with Frozen Image Encoders and Large Language Models.
\newblock In \emph{ICML}.

\bibitem[{Li et~al.(2020)Li, Xu, Cui, Huang, Wei, Li, and Zhou}]{li-etal-2020-docbank}
Li, M.; Xu, Y.; Cui, L.; Huang, S.; Wei, F.; Li, Z.; and Zhou, M. 2020.
\newblock {D}oc{B}ank: A Benchmark Dataset for Document Layout Analysis.
\newblock In \emph{COLING}, 949--960.

\bibitem[{Lin et~al.(2014)Lin, Maire, Belongie, Hays, Perona, Ramanan, Doll{\'a}r, and Zitnick}]{lin2014microsoft}
Lin, T.-Y.; Maire, M.; Belongie, S.; Hays, J.; Perona, P.; Ramanan, D.; Doll{\'a}r, P.; and Zitnick, C.~L. 2014.
\newblock Microsoft COCO: Common Objects in Context.
\newblock In \emph{ECCV}, 740--755.

\bibitem[{Liu et~al.(2023{\natexlab{a}})Liu, Li, Wu, and Lee}]{liu2023llava}
Liu, H.; Li, C.; Wu, Q.; and Lee, Y.~J. 2023{\natexlab{a}}.
\newblock Visual Instruction Tuning.
\newblock \emph{arXiv:2304.08485}.

\bibitem[{Liu et~al.(2023{\natexlab{b}})Liu, Li, Li, Yu, Huang, Peng, Liu, Chen, Li, Jin et~al.}]{liu2023hidden}
Liu, Y.; Li, Z.; Li, H.; Yu, W.; Huang, M.; Peng, D.; Liu, M.; Chen, M.; Li, C.; Jin, L.; et~al. 2023{\natexlab{b}}.
\newblock On the Hidden Mystery of OCR in Large Multimodal Models.
\newblock \emph{arXiv:2305.07895}.

\bibitem[{Longpre et~al.(2023)Longpre, Hou, Vu, Webson, Chung, Tay, Zhou, Le, Zoph, Wei et~al.}]{longpre2023flan}
Longpre, S.; Hou, L.; Vu, T.; Webson, A.; Chung, H.~W.; Tay, Y.; Zhou, D.; Le, Q.~V.; Zoph, B.; Wei, J.; et~al. 2023.
\newblock The FLAN collection: Designing Data and Methods for Effective Instruction Tuning.
\newblock \emph{arXiv:2301.13688}.

\bibitem[{Loshchilov and Hutter(2017)}]{loshchilov2017decoupled}
Loshchilov, I.; and Hutter, F. 2017.
\newblock Decoupled Weight Decay Regularization.
\newblock \emph{arXiv:1711.05101}.

\bibitem[{Lu et~al.(2022)Lu, Mishra, Xia, Qiu, Chang, Zhu, Tafjord, Clark, and Kalyan}]{lu2022learn}
Lu, P.; Mishra, S.; Xia, T.; Qiu, L.; Chang, K.-W.; Zhu, S.-C.; Tafjord, O.; Clark, P.; and Kalyan, A. 2022.
\newblock Learn to Explain: Multimodal Reasoning via Thought Chains for Science Question Answering.
\newblock In \emph{NeurIPS}.

\bibitem[{Lu et~al.(2021)Lu, Qiu, Chen, Xia, Zhao, Zhang, Yu, Liang, and Zhu}]{lu2021iconqa}
Lu, P.; Qiu, L.; Chen, J.; Xia, T.; Zhao, Y.; Zhang, W.; Yu, Z.; Liang, X.; and Zhu, S.-C. 2021.
\newblock IconQA: A New Benchmark for Abstract Diagram Understanding and Visual Language Reasoning.
\newblock In \emph{NeurIPS}.

\bibitem[{Masry et~al.(2022)Masry, Do, Tan, Joty, and Hoque}]{masry-etal-2022-chartqa}
Masry, A.; Do, X.~L.; Tan, J.~Q.; Joty, S.; and Hoque, E. 2022.
\newblock {C}hart{QA}: A Benchmark for Question Answering about Charts with Visual and Logical Reasoning.
\newblock In \emph{ACL Findings}, 2263--2279.

\bibitem[{Mathew et~al.(2022)Mathew, Bagal, Tito, Karatzas, Valveny, and Jawahar}]{Mathew_2022_WACV}
Mathew, M.; Bagal, V.; Tito, R.; Karatzas, D.; Valveny, E.; and Jawahar, C. 2022.
\newblock InfographicVQA.
\newblock In \emph{WACV}, 1697--1706.

\bibitem[{Mathew, Karatzas, and Jawahar(2021)}]{Mathew_2021_WACV}
Mathew, M.; Karatzas, D.; and Jawahar, C.~V. 2021.
\newblock DocVQA: A Dataset for VQA on Document Images.
\newblock In \emph{WACV}, 2200--2209.

\bibitem[{Mishra et~al.(2019)Mishra, Shekhar, Singh, and Chakraborty}]{MishraSSC19}
Mishra, A.; Shekhar, S.; Singh, A.~K.; and Chakraborty, A. 2019.
\newblock OCR-VQA: Visual Question Answering by Reading Text in Images.
\newblock In \emph{ICDAR}, 947--952.

\bibitem[{Mishra et~al.(2022)Mishra, Khashabi, Baral, and Hajishirzi}]{mishra-etal-2022-cross}
Mishra, S.; Khashabi, D.; Baral, C.; and Hajishirzi, H. 2022.
\newblock Cross-Task Generalization via Natural Language Crowdsourcing Instructions.
\newblock In \emph{ACL}, 3470--3487.

\bibitem[{OpenAI(2023)}]{openai2023gpt4}
OpenAI. 2023.
\newblock GPT-4 Technical Report.
\newblock \emph{arXiv:2303.08774}.

\bibitem[{Park et~al.(2019)Park, Shin, Lee, Lee, Surh, Seo, and Lee}]{park2019cord}
Park, S.; Shin, S.; Lee, B.; Lee, J.; Surh, J.; Seo, M.; and Lee, H. 2019.
\newblock CORD: A Consolidated Receipt Dataset for Post-OCR Parsing.
\newblock In \emph{Workshop on Document Intelligence at NeurIPS}.

\bibitem[{Pfitzmann et~al.(2022)Pfitzmann, Auer, Dolfi, Nassar, and Staar}]{doclaynet}
Pfitzmann, B.; Auer, C.; Dolfi, M.; Nassar, A.~S.; and Staar, P. 2022.
\newblock DocLayNet: A Large Human-Annotated Dataset for Document-Layout Segmentation.
\newblock In \emph{KDD}, 3743–3751.

\bibitem[{Radford et~al.(2021)Radford, Kim, Hallacy, Ramesh, Goh, Agarwal, Sastry, Askell, Mishkin, Clark et~al.}]{radford2021learning}
Radford, A.; Kim, J.~W.; Hallacy, C.; Ramesh, A.; Goh, G.; Agarwal, S.; Sastry, G.; Askell, A.; Mishkin, P.; Clark, J.; et~al. 2021.
\newblock Learning Transferable Visual Models from Natural Language Supervision.
\newblock In \emph{ICML}, 8748--8763.

\bibitem[{{\v{S}}imsa et~al.(2023){\v{S}}imsa, {\v{S}}ulc, U{\v{r}}i{\v{c}}{\'a}{\v{r}}, Patel, Hamdi, Koci{\'a}n, Skalick{\`y}, Matas, Doucet, Coustaty, and Karatzas}]{simsa2023docile}
{\v{S}}imsa, {\v{S}}.; {\v{S}}ulc, M.; U{\v{r}}i{\v{c}}{\'a}{\v{r}}, M.; Patel, Y.; Hamdi, A.; Koci{\'a}n, M.; Skalick{\`y}, M.; Matas, J.; Doucet, A.; Coustaty, M.; and Karatzas, D. 2023.
\newblock {DocILE} Benchmark for Document Information Localization and Extraction.
\newblock \emph{arXiv:2302.05658}.

\bibitem[{Singh et~al.(2019)Singh, Natarajan, Shah, Jiang, Chen, Batra, Parikh, and Rohrbach}]{SinghNSJCBPR19}
Singh, A.; Natarajan, V.; Shah, M.; Jiang, Y.; Chen, X.; Batra, D.; Parikh, D.; and Rohrbach, M. 2019.
\newblock Towards {VQA} Models That Can Read.
\newblock In \emph{CVPR}, 8317--8326.

\bibitem[{Sun et~al.(2021)Sun, Kuang, Yue, Lin, and Zhang}]{sun2021spatial}
Sun, H.; Kuang, Z.; Yue, X.; Lin, C.; and Zhang, W. 2021.
\newblock Spatial Dual-Modality Graph Reasoning for Key Information Extraction.
\newblock \emph{arXiv:2103.14470}.

\bibitem[{Tanaka et~al.(2023)Tanaka, Nishida, Nishida, Hasegawa, Saito, and Saito}]{SlideVQA2023}
Tanaka, R.; Nishida, K.; Nishida, K.; Hasegawa, T.; Saito, I.; and Saito, K. 2023.
\newblock SlideVQA: A Dataset for Document Visual Question Answering on Multiple Images.
\newblock In \emph{AAAI}, 13636--13645.

\bibitem[{Tanaka, Nishida, and Yoshida(2021)}]{DBLP:conf/aaai/TanakaNY21}
Tanaka, R.; Nishida, K.; and Yoshida, S. 2021.
\newblock VisualMRC: Machine Reading Comprehension on Document Images.
\newblock In \emph{AAAI}, 13878--13888.

\bibitem[{T{\"u}selmann et~al.(2022)T{\"u}selmann, M{\"u}ller, Wolf, and Fink}]{tuselmann2022recognition}
T{\"u}selmann, O.; M{\"u}ller, F.; Wolf, F.; and Fink, G.~A. 2022.
\newblock Recognition-free Question Answering on Handwritten Document Collections.
\newblock In \emph{ICFHR}, 259--273.

\bibitem[{Wang et~al.(2021)Wang, Li, Zhou, Chen, Grossman, and Li}]{wang2021screen2words}
Wang, B.; Li, G.; Zhou, X.; Chen, Z.; Grossman, T.; and Li, Y. 2021.
\newblock Screen2words: Automatic mobile UI summarization with multimodal learning.
\newblock In \emph{UIST}, 498--510.

\bibitem[{Wei et~al.(2021)Wei, Bosma, Zhao, Guu, Yu, Lester, Du, Dai, and Le}]{wei2021finetuned}
Wei, J.; Bosma, M.; Zhao, V.~Y.; Guu, K.; Yu, A.~W.; Lester, B.; Du, N.; Dai, A.~M.; and Le, Q.~V. 2021.
\newblock Finetuned language models are zero-shot learners.
\newblock In \emph{ICLR}.

\bibitem[{Xu et~al.(2020)Xu, Li, Cui, Huang, Wei, and Zhou}]{XuLCHWZ20}
Xu, Y.; Li, M.; Cui, L.; Huang, S.; Wei, F.; and Zhou, M. 2020.
\newblock LayoutLM: Pre-training of Text and Layout for Document Image Understanding.
\newblock In \emph{KDD}, 1192--1200.

\bibitem[{Xu et~al.(2021)Xu, Xu, Lv, Cui, Wei, Wang, Lu, Flor{\^{e}}ncio, Zhang, Che, Zhang, and Zhou}]{xu2020layoutlmv2}
Xu, Y.; Xu, Y.; Lv, T.; Cui, L.; Wei, F.; Wang, G.; Lu, Y.; Flor{\^{e}}ncio, D. A.~F.; Zhang, C.; Che, W.; Zhang, M.; and Zhou, L. 2021.
\newblock LayoutLMv2: Multi-modal Pre-training for Visually-rich Document Understanding.
\newblock In \emph{ACL/IJCNLP}, 2579--2591.

\bibitem[{Xu, Shen, and Huang(2023)}]{xu-etal-2023-multiinstruct}
Xu, Z.; Shen, Y.; and Huang, L. 2023.
\newblock {M}ulti{I}nstruct: Improving Multi-Modal Zero-Shot Learning via Instruction Tuning.
\newblock In \emph{ACL}, 11445--11465.

\bibitem[{Ye et~al.(2023{\natexlab{a}})Ye, Hu, Xu, Ye, Yan, Dan, Zhao, Xu, Li, Tian, Qi, Zhang, and Huang}]{ye2023mplugdocowl}
Ye, J.; Hu, A.; Xu, H.; Ye, Q.; Yan, M.; Dan, Y.; Zhao, C.; Xu, G.; Li, C.; Tian, J.; Qi, Q.; Zhang, J.; and Huang, F. 2023{\natexlab{a}}.
\newblock mPLUG-DocOwl: Modularized Multimodal Large Language Model for Document Understanding.
\newblock \emph{arXiv:2307.02499}.

\bibitem[{Ye et~al.(2023{\natexlab{b}})Ye, Xu, Xu, Ye, Yan, Zhou, Wang, Hu, Shi, Shi, Jiang, Li, Xu, Chen, Tian, Qi, Zhang, and Huang}]{ye2023mplugowl}
Ye, Q.; Xu, H.; Xu, G.; Ye, J.; Yan, M.; Zhou, Y.; Wang, J.; Hu, A.; Shi, P.; Shi, Y.; Jiang, C.; Li, C.; Xu, Y.; Chen, H.; Tian, J.; Qi, Q.; Zhang, J.; and Huang, F. 2023{\natexlab{b}}.
\newblock mPLUG-Owl: Modularization Empowers Large Language Models with Multimodality.
\newblock \emph{arXiv:2304.14178}.

\bibitem[{Zhang et~al.(2023)Zhang, Zhang, Gu, Zhou, Lipka, Yang, and Sun}]{zhang2023llavar}
Zhang, Y.; Zhang, R.; Gu, J.; Zhou, Y.; Lipka, N.; Yang, D.; and Sun, T. 2023.
\newblock LLaVAR: Enhanced Visual Instruction Tuning for Text-Rich Image Understanding.
\newblock \emph{arXiv:2306.17107}.

\bibitem[{Zhu et~al.(2023)Zhu, Chen, Shen, Li, and Elhoseiny}]{zhu2023minigpt}
Zhu, D.; Chen, J.; Shen, X.; Li, X.; and Elhoseiny, M. 2023.
\newblock MiniGPT-4: Enhancing Vision-Language Understanding with Advanced Large Language Models.
\newblock \emph{arXiv:2304.10592}.

\bibitem[{Zhu et~al.(2022)Zhu, Lei, Feng, Wang, Zhang, and Chua}]{zhu2022towards}
Zhu, F.; Lei, W.; Feng, F.; Wang, C.; Zhang, H.; and Chua, T.-S. 2022.
\newblock Towards Complex Document Understanding by Discrete Reasoning.
\newblock In \emph{ACMM}, 4857--4866.

\end{thebibliography}

\newpage
\appendix

\section{Further InstructDoc Details}

\begin{table*}[t!]
    \centering
        \scalebox{1.0}{
    \tabcolsep=3pt
    \small
    \begin{tabular}{lcccc} 
        \toprule
        Dataset & Document type & Held-out (split) & \# Held-in instances & \# Held-out instances \\ \midrule
        \textbf{DocILE}~\cite{simsa2023docile} & Invoice & & 56,369 & - \\
        \textbf{KLC}~\cite{borchmann2021due} &  Report & & 13,449 & - \\
        \textbf{Deepform}~\cite{borchmann2021due} &  Form & & 3,409 & - \\
        \textbf{FUNSD}~\cite{jaume2019funsd} & Form & \checkmark (test) & - & 2,270 \\
        \textbf{PWC}~\cite{borchmann2021due} &  Article & & 2,853 & - \\
        \textbf{Wildreceipt}~\cite{sun2021spatial} &  Receipt & & 16,188 & - \\
        \textbf{CORD}~\cite{park2019cord} &  Receipt & \checkmark (test) & - & 1,336 \\ 
        \textbf{SROIE}~\cite{huang2019icdar2019} & Receipt & & 2,503 & - \\ 
        \textbf{VisualMRC}~\cite{DBLP:conf/aaai/TanakaNY21} & Web-page &  & 21,015 & - \\
        \textbf{WebSRC}~\cite{
ChenZCJZLX021} &  Web-page & & 307,315 & -\\
        \textbf{OCRVQA}~\cite{MishraSSC19} &  Book cover & & 798,245 & -\\
        \textbf{DocVQA}~\cite{Mathew_2021_WACV} & Business document & & 39,463 & -  \\
        \textbf{HW-SQuAD}~\cite{tuselmann2022recognition} & Hand-written document & & 67,887 &-  \\ 
        \textbf{TAT-DQA}~\cite{zhu2022towards} &  Report & & 13,251 & - \\
        \textbf{WTQ}~\cite{borchmann2021due} &  Table & & 14,224 & - \\ 
        \textbf{IconQA}~\cite{lu2021iconqa} & Diagram & & 29,859 & - \\
        \textbf{AI2D}~\cite{kembhavi2016diagram} & Diagram & & 12,413 & -\\
        \textbf{ScienceQA}~\cite{lu2022learn} &  Diagram & & 12,726 & - \\
        \textbf{TextbookQA}~\cite{KembhaviSSCFH17} & Diagram & & 6,501 & -\\
        \textbf{InfographicVQA}~\cite{Mathew_2022_WACV} &  Infographic & \checkmark (dev) & -& 2,801 \\
        \textbf{ChartQA}~\cite{masry-etal-2022-chartqa} &  Chart & \checkmark (test) &- & 2,500 \\ 
        \textbf{SlideVQA}~\cite{SlideVQA2023} &  Slide deck & \checkmark (test) & -& 2,215 \\
        \textbf{DUDE}~\cite{landeghem2023document} & Open-domain & \checkmark (dev) & -& 6,315 \\ 
        \textbf{TabFact}~\cite{borchmann2021due} &  Table & \checkmark (dev) &- & 12,740 \\ 
        \textbf{LLaVAR}~\cite{zhang2023llavar} & 8 types documents & & 19,800 & -  \\ 
        \textbf{RVL-CDIP}~\cite{harley2015evaluation} &  Business document & & 320,000 & - \\ 
        \textbf{SciCap}~\cite{hsu-etal-2021-scicap-generating} & Figure & & 75,494 & - \\ 
        \textbf{Screen2Words}~\cite{wang2021screen2words} & Mobile UI & & 15,743 & -\\ 
        \textbf{DocBank}~\cite{li-etal-2020-docbank} & Article & & 3,999,811 & - \\
        \textbf{DocLaynet}~\cite{doclaynet} & Report & & 69,102 & - \\
        \bottomrule
    \end{tabular}
    }
    \caption{Datasets used in InstructDoc.}
    \label{tab:datasets}
\end{table*}

\subsection{Dataset Collection}
\subsubsection{Dataset list.}
Table~\ref{tab:datasets} shows the detail of all datasets we used in InstructDoc. It contains 5,917,602 held-in instances and 30,177 held-out instances.

\subsubsection{Query rephrasing.}
Table~\ref{tab:query} shows the detail of the query rephrasing annotation. The rephrased queries are more easily understandable phrases than the original queries.

\subsubsection{Instruction annotation.}
Table~\ref{tab:cord}-\ref{tab:doclaynet} show the examples of instructions for each task in InstructDoc. 

\subsection{Dataset Analysis}

\subsubsection{Starting words of the instructions.}

\begin{figure}[t!]
    \centering
    \includegraphics[width=.3745\textwidth]{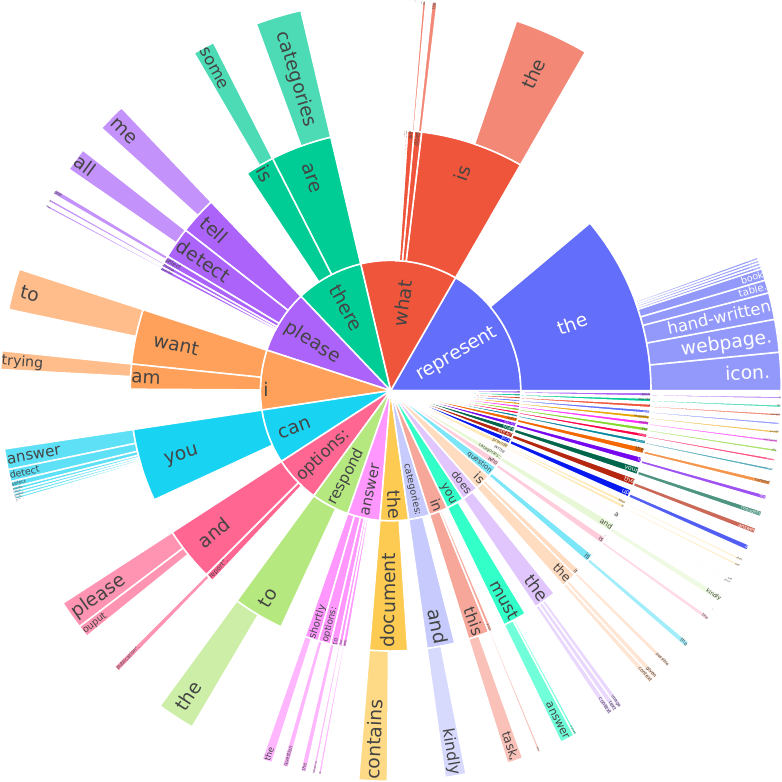}
    \caption{Distribution of first three words of the instructions.}
    \label{fig:sunburst}
\end{figure}

Figure \ref{fig:sunburst} shows the sunburst pattern of the first three words of the instructions. It can be seen that the instructions contain various types, such as questions (e.g., ``\textit{What is the}") and requests (e.g., ``\textit{I want to}") used in real-world situations.

\subsubsection{Answer styles.}
\begin{figure}[t!]
    \centering
    \includegraphics[width=.5\textwidth]{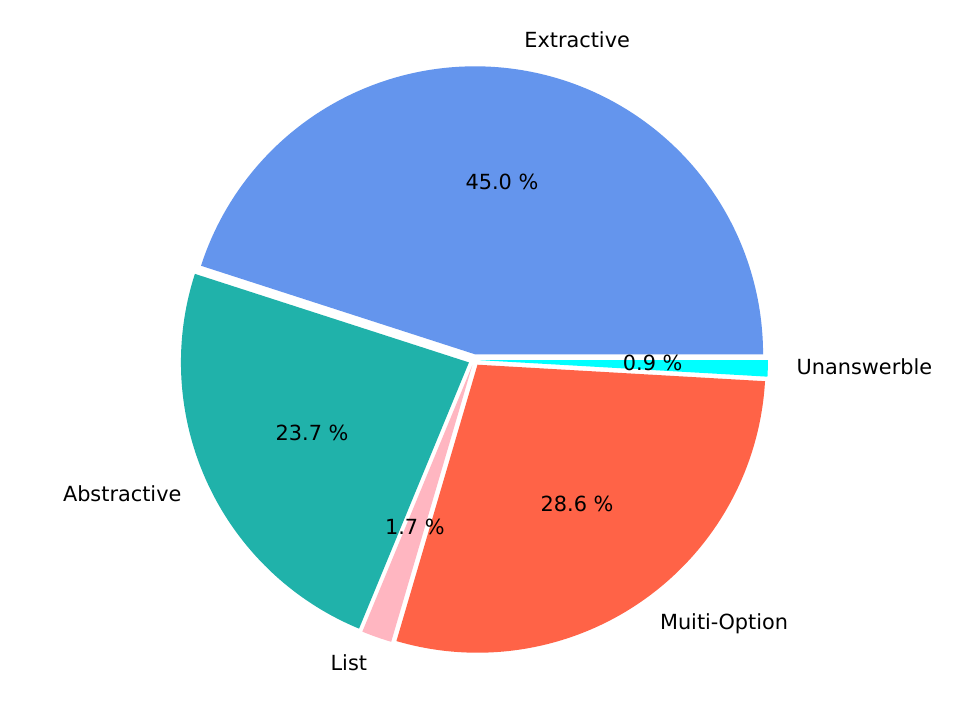}
    \caption{Distribution of the answer styles in QA datasets.}
    \label{fig:answer_types}
\end{figure}

Figure~\ref{fig:answer_types} shows InstructDoc has five diverse answer types.

\subsubsection{Word clouds.}
\begin{figure*}[t!]
    \centering
    \subfigure[Word cloud of instructions.]{
    \includegraphics[width=.31\textwidth]{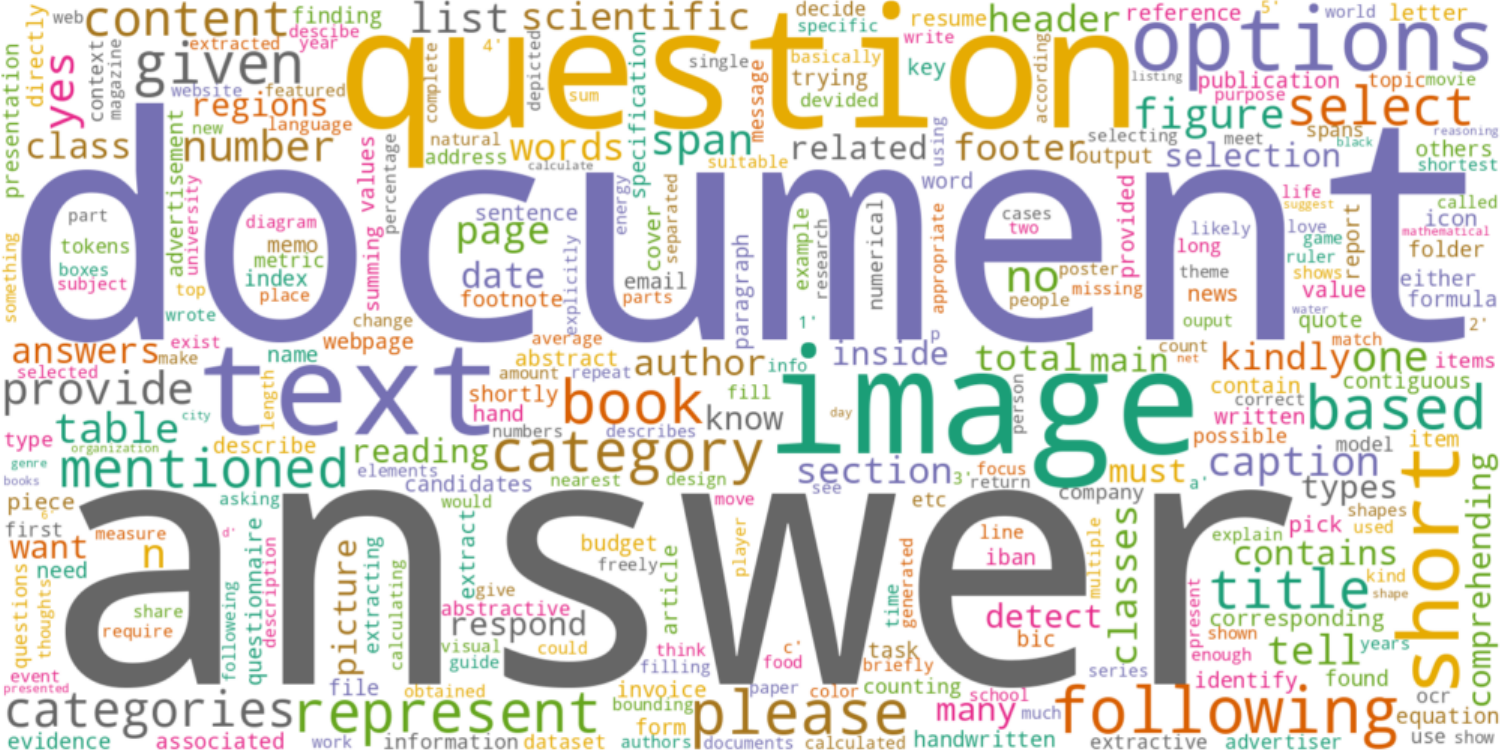}
    }
    \subfigure[Word cloud of answers.]{
    \includegraphics[width=.31\textwidth]{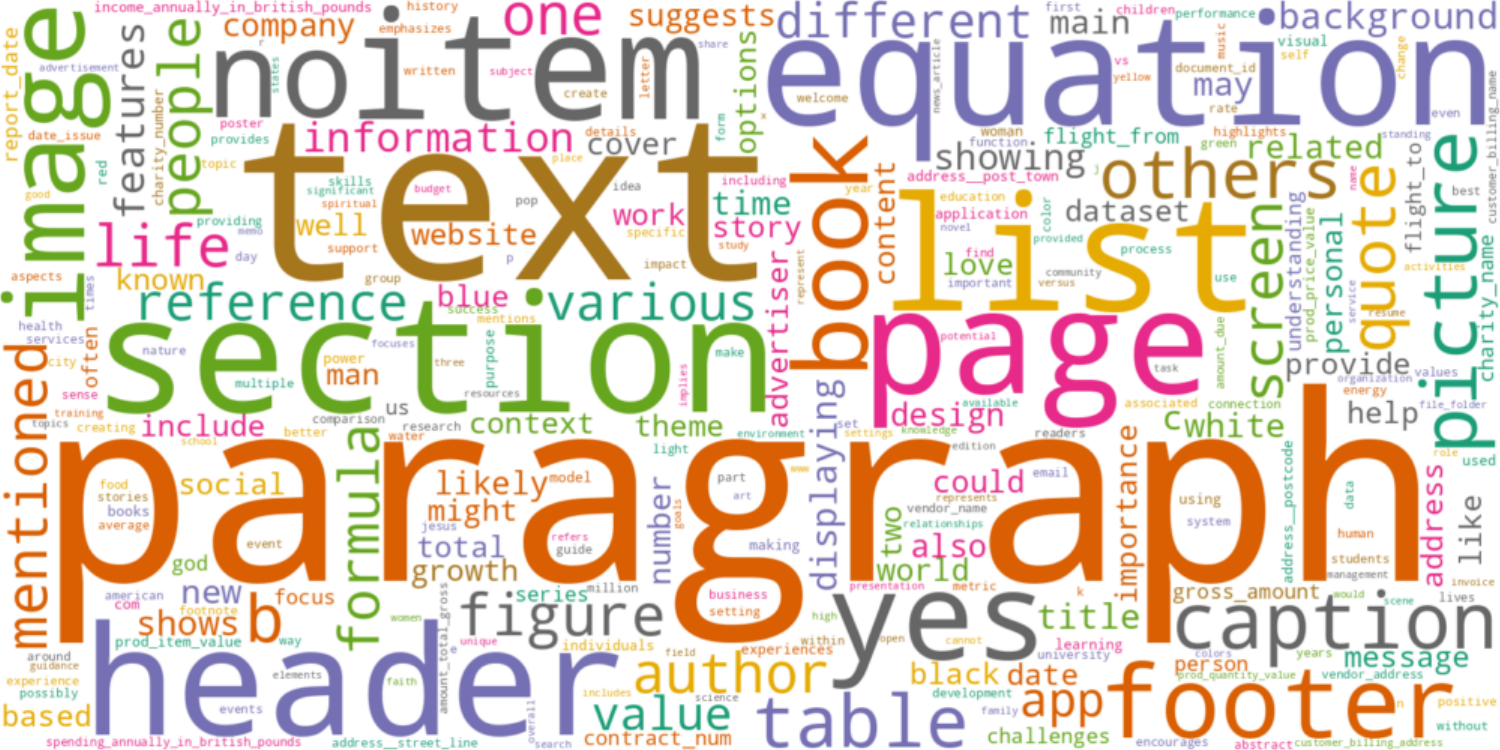}
    }
    \subfigure[Word cloud of document images.]{
    \includegraphics[width=.31\textwidth]{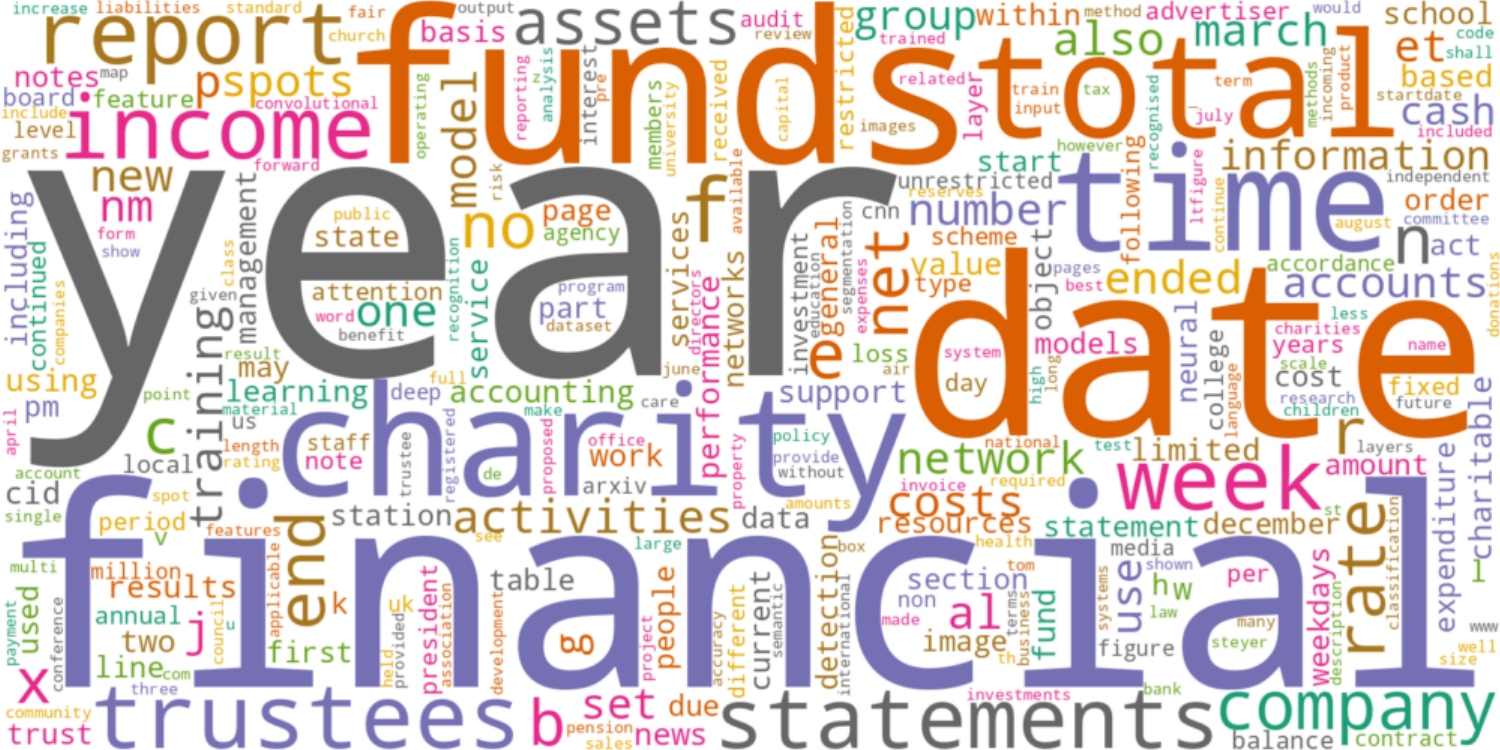}
    }
    \caption{Statistics of tokens in the instruction, answer, document images. 
    }
    \label{fig:statistics}
\end{figure*}

Figure~\ref{fig:statistics} shows how diverse the vocabulary space is in InstructDoc.

\section{Further Evaluation Setup Details}
\subsection{Main Evaluation Datasets Details}
\subsubsection{FUNSD.} Form Understanding in Noisy Scanned Documents (FUNSD)~\cite{jaume2019funsd} evaluates on the \textit{KIE} task: predicting the entity, ``title", ``key", ``value", or ``other", for the assigned text token.

\subsubsection{CORD.} Consolidated Receipt Dataset for Post-OCR Parsing (CORD)~\cite{park2019cord} is the \textit{KIE} dataset with 30 labels under 4 categories such as ``total" or ``subtotal".

\subsubsection{InfographicVQA.} This dataset focuses on the task of \textit{single-page QA w/ discrete \& visual reasoning} on infographics. It requires understanding plots/graphs, texts, and layout~\cite{Mathew_2022_WACV}.

\subsubsection{ChartQA.} This dataset focuses on the task of \textit{single-page QA w/ discrete \& visual reasoning} on chart images. We used both two subsets: (i) machine-generated set and (ii) human-written set~\cite{masry-etal-2022-chartqa}. 

\subsubsection{TabFact.} This dataset studies the task of \textit{Document NLI} with semi-structured evidence over tables. It predicts the entailment relationship between two sentences in a document~\cite{borchmann2021due}.

\subsubsection{DUDE.} Document Understanding Dataset and Evaluation (DUDE)~\cite{landeghem2023document} focuses on the task of \textit{multi-page QA w/ discrete \& visual \& multi-hop reasoning}. It is a multi-page, multi-domain, and multi-industry Document VQA for real-world document understanding. 

\subsubsection{SlideVQA.} This dataset focuses on the task of \textit{multi-page QA w/ discrete \& visual \& multi-hop reasoning} on the slide deck composed of multiple images. It requires selecting a set of evidence and answering the question~\cite{SlideVQA2023}.

\subsection{Other Evaluation Datasets Details}
\subsubsection{VisualMRC.} Visual Machine Reading Comprehension (VisualMRC)~\cite{DBLP:conf/aaai/TanakaNY21} is the task of abstractive single-page QA on the Web screenshot. We used the end-to-end setting where answers are derived from OCR results and images without ROI detection.

\subsubsection{TextVQA.} It contains scene images from Open Images dataset~\cite{kuznetsova2020open}, with questions asking to reason about text in the image~\cite{SinghNSJCBPR19}.

\subsubsection{ST-VQA.} It contains scene images from multiple sources, such as Visual Genome~\cite{KrishnaZGJHKCKL17}. We used the Open Dictionary setting where answer candidates and vocabularies are not provided at test time~\cite{BitenTMBRJVK19}.

\begin{table*}[t!]
    \centering
        \scalebox{1.0}{
    \tabcolsep=3pt
    \small
    \begin{tabular}{lc} 
        \toprule
        Dataset &  Before $\longrightarrow$ After \\ \midrule
         & \texttt{header} $\longrightarrow$ \texttt{title}\\
        FUNSD & \texttt{question} $\longrightarrow$ \texttt{key} \\
         & \texttt{answer} $\longrightarrow$ \texttt{value} \\ \midrule
         & \texttt{menu.nm} $\longrightarrow$ \texttt{menu\_name} \\
             & \texttt{menu.num} $\longrightarrow$ \texttt{menu\_id} \\
             & \texttt{menu.unitprice} $\longrightarrow$ \texttt{menu\_unitprice} \\
             & \texttt{menu.cnt} $\longrightarrow$ \texttt{menu\_quantity} \\
             & \texttt{menu.discountprice} $\longrightarrow$ \texttt{menu\_discountprice} \\
             & \texttt{menu.price} $\longrightarrow$ \texttt{menu\_price}  \\
             & \texttt{menu.itemsubtotal} $\longrightarrow$ \texttt{menu\_price\_discount\_applied}  \\
             & \texttt{menu.vatyn} $\longrightarrow$ \texttt{menu\_whether\_price\_tax\_includes}  \\
             & \texttt{menu.etc} $\longrightarrow$ \texttt{menu\_etc}  \\
             & \texttt{menu.sub\_nm} $\longrightarrow$ \texttt{submenu\_name}  \\
             & \texttt{menu.sub\_unitprice} $\longrightarrow$  \texttt{submenu\_unitprice} \\
             & \texttt{menu.sub\_cnt} $\longrightarrow$ \texttt{submenu\_quantity} \\
             & \texttt{menu.sub\_price} $\longrightarrow$ \texttt{submenu\_price} \\
             & \texttt{menu.sub\_etc} $\longrightarrow$ \texttt{submenu\_etc} \\
            CORD & \texttt{void\_menu.nm} $\longrightarrow$ \texttt{voidmenu\_name} \\
             & \texttt{void\_menu.price} $\longrightarrow$ \texttt{voidmenu\_price} \\
             & \texttt{sub\_total.subtotal\_price} $\longrightarrow$ \texttt{subtotal\_price} \\
             & \texttt{sub\_total.discount\_price} $\longrightarrow$ \texttt{subtotal\_discount\_price} \\
             & \texttt{sub\_total.service\_price} $\longrightarrow$ \texttt{subtotal\_service\_price} \\
             & \texttt{sub\_total.othersvc\_price} $\longrightarrow$ \texttt{subtotal\_chargeprice} \\
             & \texttt{sub\_total.tax\_price} $\longrightarrow$ \texttt{subtotal\_tax\_price} \\
             & \texttt{sub\_total.etc} $\longrightarrow$ \texttt{subtotal\_etc} \\
             & \texttt{total.total\_price} $\longrightarrow$ \texttt{total\_price} \\ 
             & \texttt{total.total\_etc} $\longrightarrow$ \texttt{total\_etc} \\
             & \texttt{total.cashprice} $\longrightarrow$ \texttt{total\_cashprice} \\ 
             & \texttt{total.changeprice} $\longrightarrow$ \texttt{total\_changeprice} \\
             & \texttt{total.creditcardprice} $\longrightarrow$ \texttt{total\_creditcardprice} \\ 
             & \texttt{total.emoneyprice} $\longrightarrow$ \texttt{total\_emoneyprice} \\
             & \texttt{total.menutype\_cnt} $\longrightarrow$ \texttt{total\_menutype\_count} \\
             & \texttt{total.menuqty\_cnt} $\longrightarrow$ \texttt{total\_menuquantity\_count} \\        \bottomrule
    \end{tabular}
    }
    \caption{Query rephrasing in KIE datasets. Before/After denotes the original/rephrased queries.}
    \label{tab:query}
\end{table*}

\begin{table*}[t!]
    \centering
    \tabcolsep=3pt
    \small
    \begin{tabular}{p{0.98\linewidth}} 
        \toprule
        \vspace{-1mm} \textsc{\textbf{\underline{Input}}} \\  
        \begin{minipage}{0.4\linewidth}
          \centering
          \scalebox{0.5}{\includegraphics{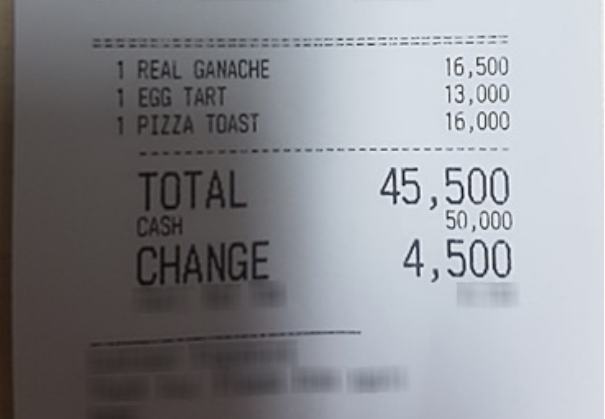}}
          \vspace{1mm} 
        \end{minipage}
        \begin{minipage}{0.6\linewidth}
        There are 30 categories for selection: \\ \\
        \text{``menu\_name``, ``menu\_id``, "menu\_unitprice", ``menu\_quantity",} \\ \text{``menu\_discountprice", ``menu\_price",``menu\_price\_discount\_applied",} \\ \text{``menu\_whether\_price\_tax\_includes", "menu\_etc", ``submenu\_name",} \\
        \text{``submenu\_unitprice", "submenu\_quantity", ``submenu\_price",``submenu\_etc",} \\
        \text{``voidmenu\_name", "voidmenu\_price", ``subtotal\_price", ``subtotal\_discount\_price",} \\
        \text{``subtotal\_service\_price", ``subtotal\_chargeprice", ``subtotal\_tax\_price",} \\
        \text{``subtotal\_etc", ``total\_price", ``total\_etc", "total\_cashprice",} \\
        \text{``total\_changeprice", ``total\_creditcardprice", ``total\_emoneyprice",} \\
        \text{"total\_menutype\_count", and ``total\_menuquantity\_count".} \\ \\
        Please output the category corresponding to the text ``16,500".
      \vspace{1mm} 
        \end{minipage}\\ \midrule
         \vspace{-1mm} \textsc{\textbf{\underline{Target}}}  \\ 
         menu\_price \\ \bottomrule
    \end{tabular}
    \caption{Example input and target for CORD dataset.}
    \label{tab:cord}
\end{table*}

\begin{table*}[t!]
    \centering
        \scalebox{1.0}{
    \tabcolsep=3pt
    \small
    \begin{tabular}{p{0.98\linewidth}} 
        \toprule
        \vspace{-1mm} \textsc{\textbf{\underline{Input}}} \\  
        \begin{minipage}{0.4\linewidth}
          \centering
          \scalebox{0.5}{\includegraphics{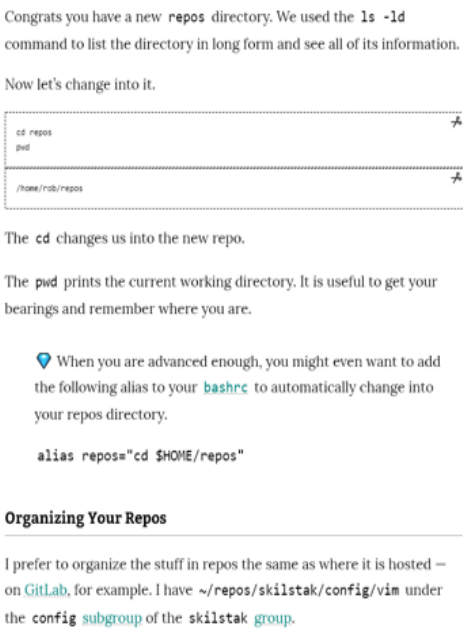}}
          \vspace{1mm} 
        \end{minipage}
        \begin{minipage}{0.58\linewidth}
Respond to the question ``What are the LS and LD commands used for?" with an abstractive answer based on the content of the document. \\ \\
You should answer with a natural language text with the correct information extracted strictly from the document.
      \vspace{1mm} 
        \end{minipage}\\ \midrule
         \vspace{-1mm} \textsc{\textbf{\underline{Target}}}  \\ 
         It is used to list the directory in long form and see all of its information.     
         \\ \bottomrule
    \end{tabular}
    }
    \caption{Example input and target for VisualMRC dataset.}
    \label{tab:visualmrc}
\end{table*}

\begin{table*}[t!]
\centering
        \scalebox{1.0}{
    \tabcolsep=3pt
    \small
    \begin{tabular}{p{0.98\linewidth}} 
        \toprule
        \vspace{-1mm} \textsc{\textbf{\underline{Input}}} \\  
        \begin{minipage}{0.4\linewidth}
          \centering
          \scalebox{0.47}{\includegraphics{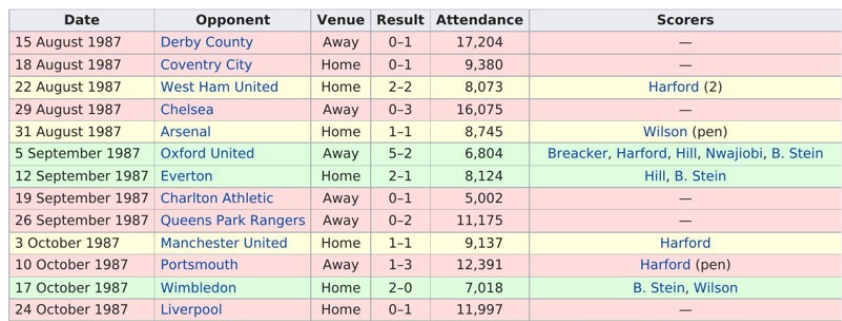}}
          \vspace{1mm} 
        \end{minipage}
        \begin{minipage}{0.58\linewidth}
You must answer to the question ``what is the difference in attendance between the first two entries?" after reading and comprehending the document. \\ \\
Answers can be either a specific span of the document or a numerical value (such as ``2," ``2.5," ``2\%," etc.). \\
Numerical questions require counting or summing values.
      \vspace{1mm} 
        \end{minipage}
        \\ \midrule
         \vspace{-1mm} \textsc{\textbf{\underline{Target}}}  \\ 
         7824     
         \\ \bottomrule
    \end{tabular}
    }
    \caption{Example input and target for WTQ dataset.}
    \label{tab:wtq}
\end{table*}

\begin{table*}[t!]
    \centering
        \scalebox{1.0}{
    \tabcolsep=3pt
    \small
    \begin{tabular}{p{0.98\linewidth}} 
        \toprule
        \vspace{-1mm} \textsc{\textbf{\underline{Input}}} \\  
        \begin{minipage}{0.4\linewidth}
          \centering
          \scalebox{0.47}{\includegraphics{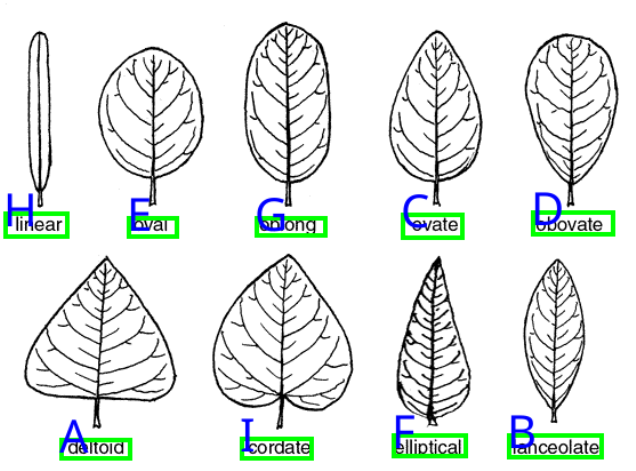}}
          \vspace{1mm} 
        \end{minipage}
        \begin{minipage}{0.58\linewidth}
Answer shortly the following question ``Which is the most narrow in this diagram?" based on the document. \\ \\

Please select one answer from the following options: \\
``Elliptical", \\
``Linear", \\
``Deltoid", \\
``Oval" \\
      \vspace{1mm} 
        \end{minipage}
        \\ \midrule
         \vspace{-1mm} \textsc{\textbf{\underline{Target}}}  \\ 
         Linear     
         \\ \bottomrule
    \end{tabular}
    }
    \caption{Example input and target for AI2D dataset.}
    \label{tab:ai2d}
\end{table*}

\begin{table*}[t!]
    \centering
        \scalebox{1.0}{
    \tabcolsep=3pt
    \small
    \begin{tabular}{p{0.98\linewidth}} 
        \toprule
        \vspace{-1mm} \textsc{\textbf{\underline{Input}}} \\  
        \begin{minipage}{0.4\linewidth}
          \centering
          \scalebox{0.5}{\includegraphics{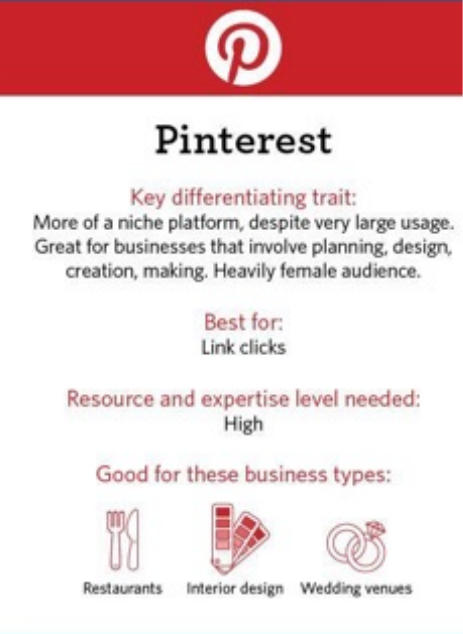}}
          \vspace{1mm} 
        \end{minipage}
        \begin{minipage}{0.58\linewidth}
There is some text on the image. \\

Provide your short answer to the question ``Which three business types is Pinterest good for?" based on the document. \\ \\
The answer type can be divided following types: \\
- Directly extracting the words from the image \\
- a list of spans inside of document (each span should be separated by ``,") \\
- not exist explicitly as a span of document \\ 
      \vspace{1mm} 
        \end{minipage}
        \\ \midrule
         \vspace{-1mm} \textsc{\textbf{\underline{Target}}}  \\ 
         Interior design, Restaurants, Wedding venues
         \\ \bottomrule
    \end{tabular}
    }
    \caption{Example input and target for InfographicVQA dataset.}
    \label{tab:infovqa}
\end{table*}

\begin{table*}[t!]
    \centering
        \scalebox{1.0}{
    \tabcolsep=3pt
    \small
    \begin{tabular}{p{0.98\linewidth}} 
        \toprule
        \vspace{-1mm} \textsc{\textbf{\underline{Input}}} \\  
        \begin{minipage}{0.4\linewidth}
          \centering
          \scalebox{0.5}{\includegraphics{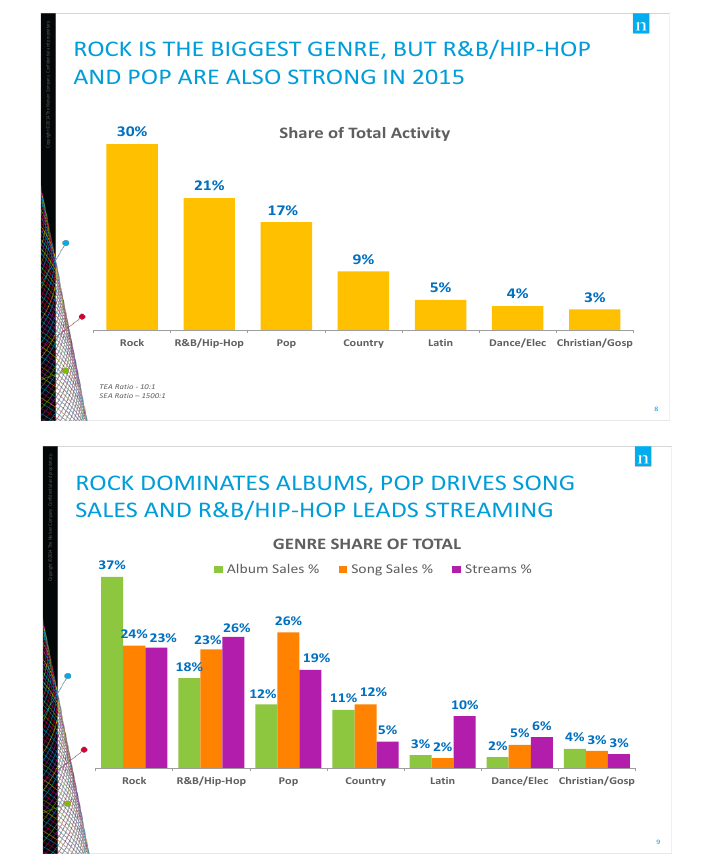}}
          \vspace{1mm} 
        \end{minipage}
        \begin{minipage}{0.58\linewidth}
Respond to the question ``What is the combined percentage of Album Sales \% and Song Sales \% for the genre with a 9\% Share of Total Activity?" with a short answer based on the content of the multi-page document. \\ \\
If you need mathematical reasoning, you first extract numbers and then calculate them to answer the question. \\
Answers contain either: \\
- a span inside of document \\
- a list of spans inside of document (each span should be separated by ``,") \\
- not exist explicitly as a span of document (the answer should be freely generated text) \\ \\
Directly answer the question from the document with 1 to 3 words as possible.
      \vspace{1mm} 
        \end{minipage}
        \\ \midrule
         \vspace{-1mm} \textsc{\textbf{\underline{Target}}}  \\ 
         23\%
         \\ \bottomrule
    \end{tabular}
    }
    \caption{Example input and target for SlideVQA dataset.}
    \label{tab:slidevqa}
\end{table*}

\begin{table*}[t!]
    \centering
        \scalebox{1.0}{
    \tabcolsep=3pt
    \small
    \begin{tabular}{p{0.98\linewidth}} 
        \toprule
        \vspace{-1mm} \textsc{\textbf{\underline{Input}}} \\  
        \begin{minipage}{0.4\linewidth}
          \centering
          \scalebox{0.5}{\includegraphics{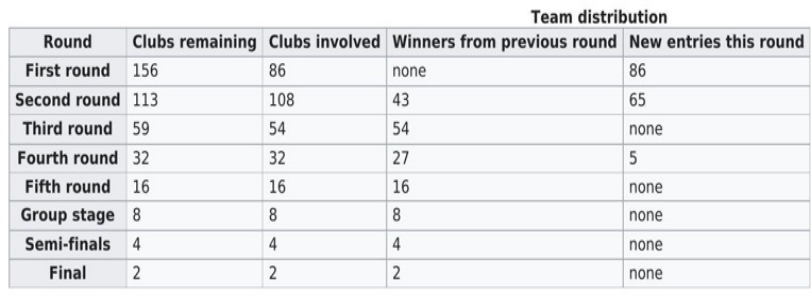}}
          \vspace{1mm} 
        \end{minipage}
        \begin{minipage}{0.58\linewidth}
        I want to know a short answer to the question. \\ \\
        Can we conclude that ``during the third round of the turkish cup , there be no new entry during that stage", based on the content of the given document? \\ \\
        Answers are selected from given options ``yes" or ``no", and you should not output other than the selected option.
      \vspace{1mm} 
        \end{minipage}
         \\ \midrule
         \vspace{-1mm} \textsc{\textbf{\underline{Target}}}  \\ 
         yes \\ \bottomrule
    \end{tabular}
    }
    \caption{Example input and target for TabFact dataset.}
    \label{tab:tabfact}
\end{table*}

\begin{table*}[t!]
    \centering
        \scalebox{1.0}{
    \tabcolsep=3pt
    \small
    \begin{tabular}{p{0.98\linewidth}} 
        \toprule
        \vspace{-1mm} \textsc{\textbf{\underline{Input}}} \\  
        \begin{minipage}{0.4\linewidth}
          \centering
          \scalebox{0.5}{\includegraphics{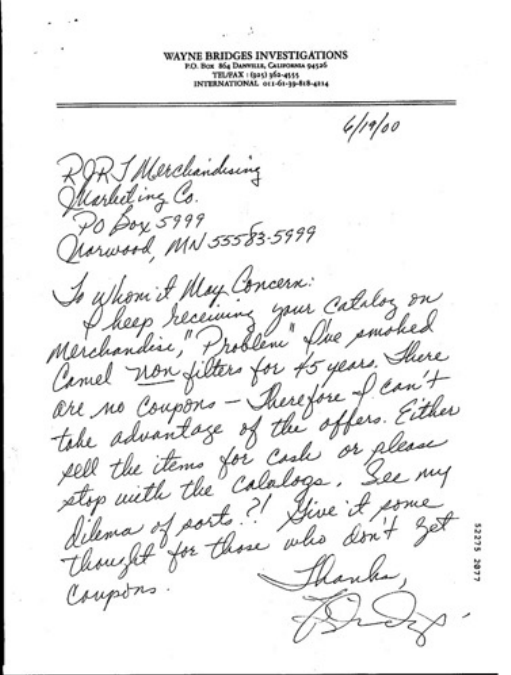}}
          \vspace{1mm} 
        \end{minipage}
        \begin{minipage}{0.58\linewidth}
Pick up a suitable class from following classes: \\ \\ 

``letter", ``form", ``email", ``handwritten", ``advertisement", ``scientific report", ``scientific publication", ``specification", ``file folder", ``news article", ``budget", ``invoice", "presentation", ``questionnaire", ``resume", and ``memo".
      \vspace{1mm} 
        \end{minipage}
        \\ \midrule
         \vspace{-1mm} \textsc{\textbf{\underline{Target}}}  \\ 
         handwritten \\ \bottomrule
    \end{tabular}
    }
    \caption{Example input and target for RVL-CDIP dataset.}
    \label{tab:rvlcdip}
\end{table*}

\begin{table*}[t!]
    \centering
        \scalebox{1.0}{
    \tabcolsep=3pt
    \small
    \begin{tabular}{p{0.98\linewidth}} 
        \toprule
        \vspace{-1mm} \textsc{\textbf{\underline{Input}}} \\  
        \begin{minipage}{0.4\linewidth}
          \centering
          \scalebox{0.5}{\includegraphics{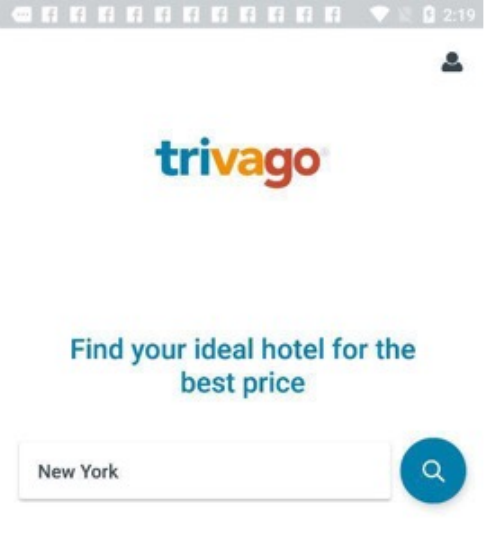}}
          \vspace{1mm} 
        \end{minipage}
        \begin{minipage}{0.58\linewidth}
Share your thoughts by using a few words on the content of the image.
      \vspace{1mm} 
        \end{minipage}
        \\ \midrule
         \vspace{-1mm} \textsc{\textbf{\underline{Target}}}  \\ 
Page with search bar to search hotel.
         \\ \bottomrule
    \end{tabular}
    }
    \caption{Example input and target for Screen2Words dataset.}
    \label{tab:screen2words}
\end{table*}

\begin{table*}[t!]
    \centering
        \scalebox{1.0}{
    \tabcolsep=3pt
    \small
    \begin{tabular}{p{0.98\linewidth}} 
        \toprule
        \vspace{-1mm} \textsc{\textbf{\underline{Input}}} \\  
        \begin{minipage}{0.4\linewidth}
          \centering
          \scalebox{0.4}{\includegraphics{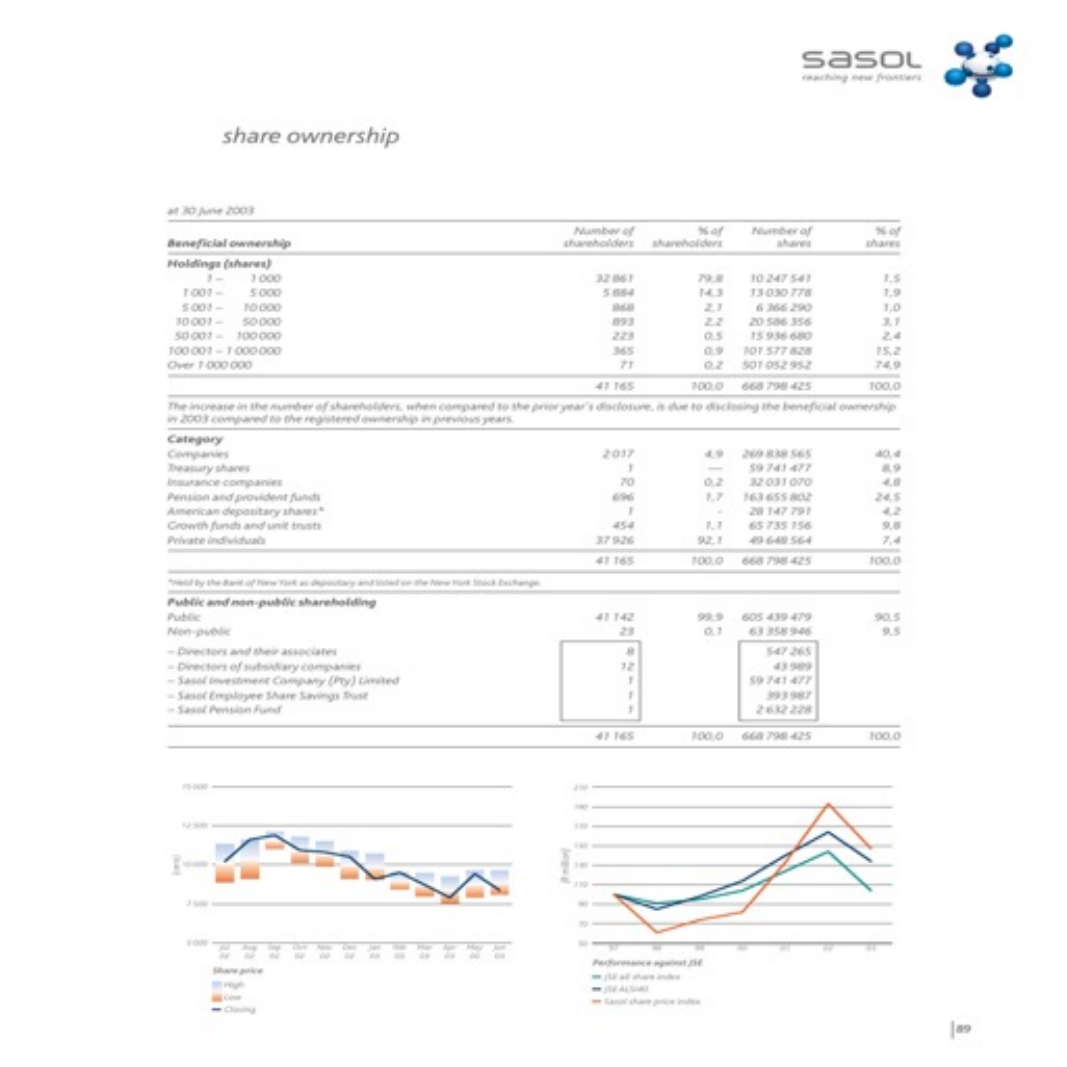}}
          \vspace{1mm} 
        \end{minipage}
        \begin{minipage}{0.58\linewidth}
Please detect all regions that meet following classes: \\ \\
``Abstract", ``Author", ``Caption", ``Date", ``Equation", ``Figure", "Footer", "List",  ``Paragraph", ``Reference", ``Section", ``Table", and ``Title".
      \vspace{1mm} 
        \end{minipage}
        \\ \midrule
         \vspace{-1mm} \textsc{\textbf{\underline{Target}}}  \\ 
Picture [750, 29, 954, 94] Section-header [209, 120, 376, 139] Table [151, 191, 850, 702] Picture [521, 731, 837, 946] Picture [158, 733, 485, 955] Page-footer [897, 962, 911, 972]
         \\ \bottomrule
    \end{tabular}
    }
    \caption{Example input and target for DocLaynet dataset.}
    \label{tab:doclaynet}
\end{table*}

\end{document}